\documentclass[10pt,journal,compsoc]{IEEEtran}
\ifCLASSOPTIONcompsoc
\else
\fi
\ifCLASSINFOpdf
\else
\fi
\usepackage{hyperref}
\usepackage{forest}
\usepackage[noadjust]{cite}
\usepackage{multirow}
\usepackage{amsmath}
\usepackage{pifont}
\newcommand{\cmark}{\ding{51}}%
\newcommand{\xmark}{\ding{55}}%
\usepackage{caption}
\begin{document}

%
\title{AMMUS : A Survey of Transformer-based Pretrained Models in Natural Language Processing}
%
%
%
%

\author{Katikapalli~Subramanyam~Kalyan,
        Ajit~Rajasekharan,
        and~Sivanesan~Sangeetha
\IEEEcompsocitemizethanks{\IEEEcompsocthanksitem K.S.Kalyan is with the Department
of Computer Applications, National Institute of Technology Trichy, Trichy,
Tamil Nadu, India, 620015.\protect\\
E-mail: kalyan.ks@yahoo.com, Website: \url{https://mr-nlp.github.io}
\IEEEcompsocthanksitem Ajit Rajasekharan is with the Nference.ai as CTO, Cambridge, MA, USA, 02142.
\IEEEcompsocthanksitem S.Sangeetha  is with the Department of Computer Applications, National Institute of Technology Trichy, Trichy, Tamil Nadu, India, 620015..}
\thanks{Preprint under review - The paper is named (AMMUS - AMMU Smiles) in the memory of one of the close friends of K.S.Kalyan (\url{https://mr-nlp.github.io}).}}

\IEEEtitleabstractindextext{%
\begin{abstract}

Transformer-based pretrained language models (T-PTLMs) have achieved great success in almost every NLP task. The evolution of these models started with GPT and BERT. These models are built on the top of transformers, self-supervised learning and transfer learning. Transformed-based PTLMs learn universal language representations from large volumes of text data using self-supervised learning and transfer this knowledge to downstream tasks. These models provide good background knowledge to downstream tasks which avoids training of downstream models from scratch.  In this comprehensive survey paper, we initially give a brief overview of self-supervised learning. Next, we explain various core concepts like pretraining, pretraining methods, pretraining tasks, embeddings and downstream adaptation methods. Next, we present a new taxonomy of T-PTLMs and then give brief overview of various benchmarks including both intrinsic and extrinsic. We present a summary of various useful libraries to work with T-PTLMs. Finally, we highlight some of the future research directions which will further improve these models.  We strongly believe that this comprehensive survey paper will serve as a good reference to learn the core concepts as well as to stay updated with the recent happenings in T-PTLMs. The list of T-PTLMs along with links is available at \underline{\textbf{\url{https://mr-nlp.github.io/posts/2021/05/tptlms-list/}}}
\end{abstract}

}

\maketitle

\IEEEdisplaynontitleabstractindextext

%
\IEEEpeerreviewmaketitle

\tableofcontents
\section{Introduction}
\IEEEPARstart{T}{ransformer}-based pretrained language models (T-PTLMs) like GPT-1 \cite{radford2018improving}, BERT \cite{devlin2019bert}, XLNet \cite{yang2019xlnet}, RoBERTa \cite{liu2019roberta}, ELECTRA \cite{clark2019electra}, T5 \cite{raffel2019exploring}, ALBERT \cite{lan2019albert},  BART \cite{lewis2020bart} and PEGAUSUS \cite{zhang2020pegasus} have achieved tremendous success in NLP because of their ability to learn universal language representations from large volumes of unlabeled text data and then transfer this knowledge to downstream tasks. In the early days, NLP systems are mostly rule-based which are later replaced by machine-learned models. Machine learning models require feature engineering which requires domain expertise and it is a time-consuming process too. The evolution of better computer hardware like GPUs and word embeddings like Word2Vec \cite{mikolov2013efficient} and Glove \cite{pennington2014glove} increased the use of deep learning models like CNN \cite{kalchbrenner2014convolutional} and RNN \cite{liu2016recurrent,zhou2016text} for building NLP systems. The main drawback with these deep learning models is the requirement of training the model from scratch except for the word embeddings. Training the model from scratch requires a large number of labeled instances which are expensive to generate. However, we expect the model to perform well using few labeled instances only. Transfer learning \cite{pan2009survey} allows the reuse of knowledge learned in source tasks to perform well in the target task. Here the target task should be similar to the source task. Based on the idea of transfer learning, researchers in Computer Vision trained large CNN models \cite{simonyan2014very,szegedy2016rethinking,he2016deep,tan2019efficientnet} using large scale labeled datasets like ImageNet \cite{krizhevsky2012imagenet,russakovsky2015imagenet}. These models learn image representations which are common across all the tasks. The large pretrained CNN models are adapted to downstream tasks by including few task-specific layers and then fine-tuned on the target datasets \cite{kaur2019automated}. As the pretrained CNN models provide good background knowledge to the downstream models, they enjoyed tremendous success in many CV tasks \cite{he2016deep,ren2015faster}.

Deep learning models like CNN and RNN have difficulties in modelling long term contexts and learn the word representations with locality bias \cite{qiu2020pre}. Moreover, as RNNs process the input sequentially i.e., word by word, the utilization of parallel computer hardware is limited. To overcome these drawbacks in existing deep learning models, Vaswani et al. \cite{vaswani2017attention} proposed a deep learning model called Transformers which is completely based on self-attention.  Self-attention allows for more parallelization compared to RNNs and can easily model long term contexts as every token attend to all the tokens in the input sequence \cite{vaswani2017attention}. Transformers contains a stack of encoder and decoder layers. With the help of a stack of encoder and decoder layers, transformers can learn complex language information. It is a very expensive and time-taking process to generate a large amount of labeled data in the NLP domain. However, it is very easy to get large volumes of unlabeled text data. NLP research community impressed with the success of CNN-based pretrained models in Computer Vision, have developed T-PTLMs by combining the power of transformers and self-supervised learning. Self-supervised learning allows the transformers to learn based on the pseudo supervision provided by one or more pretraining tasks. 

GPT and BERT are the first T-PTLMs developed based on transformer decoder and encoder layers respectively. Following GPT and BERT, models like XLNet , RoBERTa, ELECTRA, ALBERT, T5, BART and PEGAUSUS are proposed. Here XLNet, RoBERTa, ELECTRA and ALBERT are improvements over BERT model while T5, BART and PEGAUSUS are encoder-decoder based models. Kaplan et al. \cite{kaplan2020scaling} showed that the performance of T-PTLMs can be increased just by increasing the size of the model. This observation triggered the development of large-scale T-PTLMs like GPT-3 (175B) \cite{brown2020language}, PANGU- (200B) \cite{zeng2021pangu}, GShard (600B) \cite{lepikhin2020gshard} which contain billions of parameters and Switch-Transformers (1.6T) \cite{fedus2021switch} which contains trillions of parameters. Following the success of T-PTLMs in general English domain, T-PTLMs are also developed for other domains like Finance \cite{yang2020finbert}, Legal \cite{leivaditi2020benchmark,chalkidis2020legal}, News \cite{gururangan2020don}, Programming \cite{lu2021codexglue,feng2020codebert,ahmad2021unified,guo2020graphcodebert,phan2021cotext}, Dialogue \cite{wu2020tod}, Networking \cite{louis2020netbert}, Academic \cite{liu2021oag,beltagy2019scibert,peng2021mathbert}  and Biomedical  \cite{lee2020biobert,alsentzer2019publicly,gu2020domain,peng2019transfer}.  T-PTLMs support transfer-learning also as these models can be adapted to downstream tasks by fine-tuning or prompt-tuning on target datasets. In this survey paper, we present a comprehensive review of recent research works related to T-PTLMs. We summarize the highlights of our survey as
\begin{itemize}
    \item We present a brief overview of SSL, the backbone behind developing T-PTLMs (Section \ref{ssl-sec}).
    \item We explain various core concepts related to T-PTLMs like pretraining, pretraining methods, pretraining tasks, embeddings and downstream adaptation methods (Section \ref{core-concepts-sec}).
    \item We present a new taxonomy to categorize various T-PTLMs. This taxonomy is based on four perspectives namely pretraining corpus, architecture, type of SSL and extensions (Section \ref{taxonomy-sec}). 
    \item We present a new taxonomy to categorize various downstream adaptation methods and explain each in detail (Section \ref{adaptation-methods-sec}).
   \item We present a brief overview of various benchmarks including both intrinsic and extrinsic which evaluate the progress of T-PTLMs (Section \ref{evaluation}).
    \item We present a brief overview of various libraries starting from Huggingface Transformers to Transformer-interpret which are useful to work T-PTLMs (Section \ref{libraries-sec}).
    \item  We briefly discuss some of the future research directions which will drive the research community to further improve the models (Section \ref{future-directions}).
\end{itemize}

\section{Self-Supervised Learning (SSL) }
\label{ssl-sec}
Self-supervised learning, a relatively new learning paradigm has gained attention in the Artificial Intelligence (AI) research community due to its ability to make use of unlabeled data to inject universal knowledge about language, image or speech into pretrained models. Due to its data efficiency and generalization ability, SSL finds applications in various AI fields like Robotics \cite{liu2020self}, Speech  \cite{baevski2020wav2vec,sivaraman2020self}, Natural Language Processing \cite{qiu2020pre,liu2020survey} and Computer Vision  \cite{khan2021transformers,han2020survey}.  
 \subsection{Why Self-Supervised Learning?}
Supervised learning has played a crucial part in AI progress by allowing the models to learn from human-annotated instances. Models trained using supervised learning over labeled instances perform well on a specific task. However, a model trained using supervised learning requires a large number of labeled instances to achieve good performance. Data collection and labelling is a time-taking and expensive process. Moreover, it is difficult to obtain labeled data in specific domains like Medical and Legal. Further, the model learns only what is available in the training data and suffers from generalization error and spurious correlations. Although supervised learning is a dominant learning paradigm in developing AI models in the last two decades, the bottlenecks in supervised learning have forced the research community to look for alternative learning paradigms like Self-Supervised Learning (SSL). SSL does not require human labeled data and helps the model to gain more generalization ability by learning from large amounts of unlabeled data.  We summarize the drawbacks of supervised learning as
\begin{itemize}
    \item heavy dependence on human labeled instances which are expensive and time-consuming to generate.
    \item lack of generalization ability and suffers from spurious correlations.
    \item many domains like Medical and Legal are labeled data starved which limits the application of AI models in these domains.
    \item inability to learn from large amount of freely available unlabeled data.
\end{itemize}

\subsection{What is Self-Supervised Learning?}
Self-Supervised Learning (SSL) is a new learning paradigm which helps the model to learn universal knowledge based on the pseudo supervision provided by pretraining tasks. In SSL, the labels are automatically generated based on data attributes and the definition of pretraining task. Let $X = {(x_1, p_1) , (x_2, p_2), (x_3,p_3),…,(x_n,p_n)}$ represents pseudo labeled instances. The pretraining loss ($L_{SSL}$) of SSL learning paradigm can be defined as
\begin{equation}
    \displaystyle L_{SSL} = \lambda_1L_{PT-1}+\lambda_2L_{PT-2}+...+\lambda_mL_{PT-m} 
\end{equation}
Here $L_{PT-1}()$, $L_{PT-2}()$,…,$L_{PT-m}$  represent the loss functions of ‘m’ pretraining tasks and $\lambda_1()$, $\lambda_2()$,…,$\lambda_m()$ represents weights. In general, pretraining using SSL paradigm can involve more than one pretraining task. For example, RoBERTa is pretrained using only masked language modelling (MLM)  while BERT model is pretrained using two pretraining tasks namely masked language modelling (MLM) and next sentence prediction (NSP).   In case of MLM, the loss function used is cross entropy loss and in case of NSP, it is sigmoid loss.  By solving the pretraining tasks over vast amount of unlabeled data, the model learns general language representations which can encode both syntax and semantic information. These representations are useful in downstream tasks and helps the model to achieve much better performance using few labeled instances only. We can say that pretraining over vast amount of unlabeled data using SSL helps the model to gain basic common sense or background knowledge without which the model requires more labeled instances to achieve a good performance. 

SSL has similarities with other popular learning paradigms like supervised and unsupervised learning. SSL is like unsupervised learning as it does not require human labeled instances. However, it is different from unsupervised learning because a) SSL requires supervision unlike unsupervised learning and b) the objective of unsupervised learning is to identify the hidden patterns while the objective of SSL is to learn meaningful representations. SSL is like supervised learning as both the learning paradigms require supervision. However, it is different from supervised learning because a) SSL generates labels automatically without any human involvement and b) the goal of supervised learning is provide task specific knowledge while SSL aims to provide the model with universal knowledge. We summarize the goals of SSL as
\begin{itemize}
    \item learn universal language representations which provides a good background to the downstream model.
    \item better generalization ability by learning over vast amount of freely available unlabeled text data.
\end{itemize}

\subsection{Types of Self-Supervised Learning} 
Self-Supervised Learning can be classified into Generative SSL, Contrastive SSL and Adversarial SSL . Generative SSL allows the model to learn by decoding the encoded input. Generative SSL can use autoregressive, autoencoding or hybrid language models. Autoregressive language model predicts the next tokens based on the previous tokens. GPT-1 \cite{radford2018improving} is the first PTLM that is based on the autoregressive language model. Autoencoding language model predicts the masked tokens based on the unmasked tokens (bidirectional context). For example, masked language modelling (MLM) involves two steps. The first step is to encode the masked tokens using bidirectional context and the second step is to decode (predict) the original tokens based on the encoded masked token representations. Models like BERT \cite{devlin2019bert}, RoBERTa \cite{liu2019roberta} and ALBERT \cite{lan2019albert} are pretrained using MLM. Hybrid language models combine the advantages of autoregressive and autoencoding language models. For example, permutation language modelling (PLM) in XLNet \cite{yang2019xlnet}  is an example of a hybrid language model. 

Contrastive SSL allows the model to learn by comparing. Next sentence prediction (NSP) in BERT  and sentence order prediction in ALBERT  are examples of contrastive SSL. NSP involves identifying whether the given sentence pair includes consecutive sentences or not, while SOP involves identifying whether the given pair includes swapped sentences or not. Adversarial SSL allows the model to learn by identifying whether the tokens in the input sentence are replaced or shuffled or randomly substituted. Replaced token detection (RTD) in ELECTRA \cite{clark2019electra}, shuffled token detection (STD) \cite{panda2021shuffled} and random token substitution (RTS) \cite{di2021efficient} are examples of Adversarial SSL. For detailed information about SSL and types, please refer to the survey paper on SSL \cite{liu2020self}.

\section{T-PTLM Core Concepts}
\label{core-concepts-sec}
%
%

%
%
%
%
\subsection{Pretraining}
Pretraining on large volumes of unlabeled text and then fine-tuning on small task-specific datasets has become a standard approach in modern natural language processing. In Computer Vision, large models \cite{simonyan2014very,szegedy2016rethinking,he2016deep,tan2019efficientnet}  based on CNN are pretrained on large, labeled datasets like ImageNet \cite{krizhevsky2012imagenet,russakovsky2015imagenet}, and then these models are used in similar target tasks by adding few task-specific layers \cite{kaur2019automated}. Here pretraining allows the model to learn common image features which are useful in many tasks. Inspired by the success of pretrained image models, NLP researchers developed models like BERT \cite{devlin2019bert}, RoBERTa \cite{liu2019roberta}, ELECTRA \cite{clark2019electra}, XLNet \cite{yang2019xlnet}, and T5 \cite{raffel2019exploring} by pretraining them on large volumes of unlabelled text using self-supervised learning. Some of the benefits of pretraining are
\begin{itemize}
\item It helps the model to learn universal language representations by leveraging large volumes of unlabeled text.
\item Pretrained models can be adapted to downstream tasks by just adding one or two specific layers. Hence it avoids training the downstream model (except task-specific layers) from scratch by providing a good initialization.
\item It helps the model to perform better even with small datasets and hence reduces the requirement of a large number of labeled instances. 
\item Deep learning models due to having a large number of parameters tend to overfit on small datasets. As pretraining provides a good initialization, it avoids overfitting on small datasets, and hence pretraining can be viewed as a form of regularization \cite{erhan2010does}. 
\end{itemize}

\subsubsection{Pretraining Steps}
Pretraining a model involves the following five steps

\begin{figure*}[t]
\begin{center}
\includegraphics[width=18cm, height=7cm]{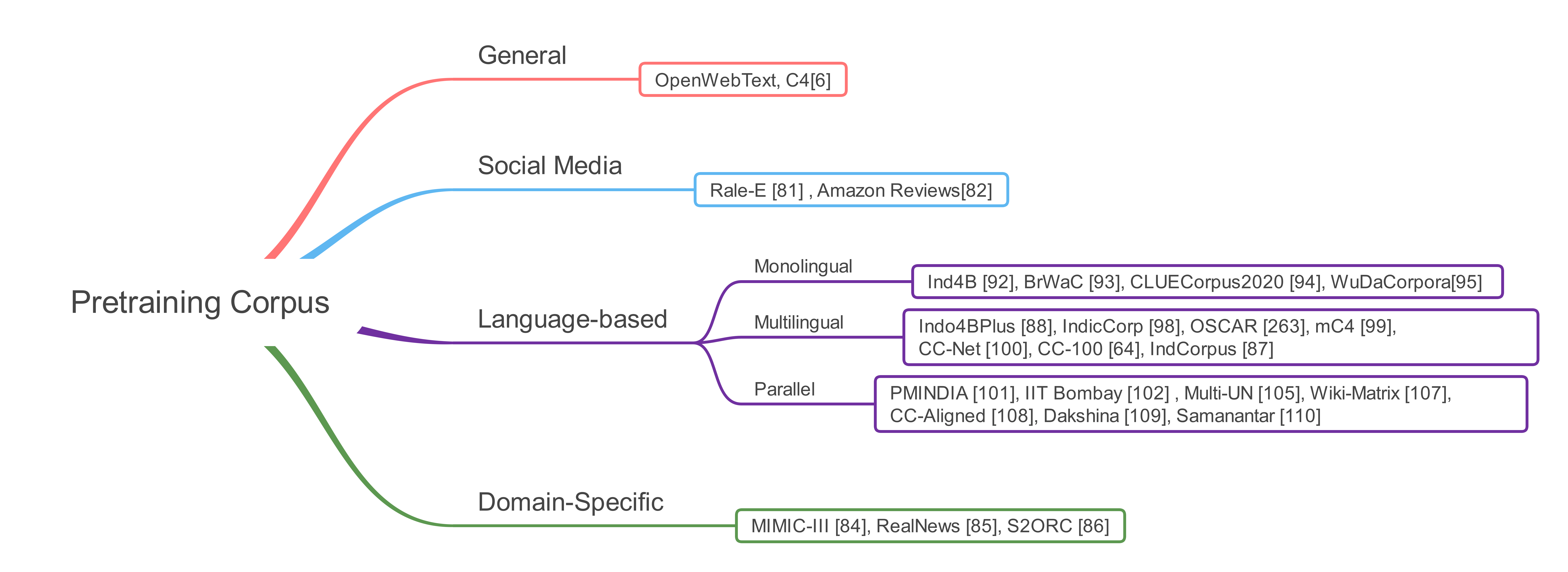}
\caption{\label{pretraining-corpora} Pretraining corpus} 
\end{center}
\end{figure*}

\textit{1. Prepare the pretraining corpus} – Pretraining corpus is obtained from one or more sources of unlabelled text and then cleaned. BERT \cite{devlin2019bert} model is pretrained on English Wikipedia and BooksCorpus. Further research \cite{liu2019roberta,raffel2019exploring,yang2019xlnet} showed that pretraining the model on a much larger text corpus obtained from multiple sources further improves the performance of the model. Moreover, Lee et al. \cite{lee2021deduplicating} showed there is a lot of redundancy in pretraining corpus in the form of near-duplicate sentences and long repetitive substrings. Further, Lee et al. \cite{lee2021deduplicating} showed pretraining the model on deduplicated corpus requires fewer training steps to achieve similar performance. 

\textit{2. Generate the vocabulary} – Most of the transformer-based pretrained language models use tokenizers like WordPiece \cite{wu2016google}, Byte Pair Encoding (BPE) \cite{sennrich2016neural}, Byte Level BPE (bBPE) \cite{radford2019language}, and SentencePiece \cite{kudo2018sentencepiece} to generate the vocabulary. Usually, vocabulary consists of all the unique characters and commonly used subwords and words. Vocabulary is generated by applying any of the tokenizers on the pretraining corpus. Different T-PTLMs use different tokenizers and generate vocabulary with different sizes. For example, BERT uses WordPiece vocabulary of size around 30K, RoBERTa uses bBPE vocabulary of size around 50K, XLM \cite{lample2019cross} uses BPE vocabulary of size 95K, mBERT \cite{devlin2019bert} WordPiece vocabulary of size 110K,  XLM-R \cite{conneau2020unsupervised}, and  mBART \cite{liu2020multilingual} uses SentencePiece vocabulary of size 250K.  The large vocabulary size in multilingual models like XLM, XLM-R, mBERT, and mBART make sense as they have to represent multiple languages. However, the size of the pretrained model increases with an increase in vocabulary size. This step is optional in the case of char-based T-PTLM like CharacterBERT \cite{el2020characterbert} and tokenization-free T-PTLMs like CANINE \cite{clark2021canine}, ByT5 \cite{xue2021byt5}, and Charformer \cite{tay2021charformer}. 

\textit{3. Design the pretraining tasks} -  During pretraining, the model learns language representations by minimizing losses based on one or more pretraining tasks. A pretraining task should 

\begin{itemize}
    \item \textit{be challenging enough to allow the model to learn semantics at word, phrase, sentence, or document level}. For example, recent research works \cite{liu2019roberta,lan2019albert} questioned the efficiency of NSP task and resulted in new pre-training tasks to learn semantics at sentence level like sentence order prediction \cite{lan2019albert} and sentence structure prediction \cite{wang2019structbert}.
    \item \textit{provide more training signal so that the model learns more language information with less pretraining corpus}. For example, RTD provides more training signal compared to MLM because RTD is defined over all the input tokens while MLM is defined over a subset of tokens only \cite{clark2019electra}.
    \item \textit{close to downstream tasks}. For example, span boundary pretraining task in SpanBERT \cite{joshi2020spanbert} is close to the span extraction task and the gap sentence generation in PEGAUSUS \cite{zhang2020pegasus} is close to the summarization task. Recent research works resulted in better versions of MLM like Swapped Language Modeling \cite{di2021efficient} which avoids the use of special mask tokens and hence reduces the discrepancy between pretraining and fine-tuning.
\end{itemize}

\textit{4. Choose the pretraining method} – Training a new model from scratch using SSL only is highly expensive and consumes a lot of pretraining time. Instead of training from scratch using SSL only, pretraining methods like KIPT \cite{qin2021knowledge,zhang2021cpm} which pretrain a model using both SSL and KD can be used. In the case of adapting general models to specific domains, pretraining methods like continual pretraining with new vocabulary \cite{kuratov2019adaptation,souza2020bertimbau,arkhipov2019tuning,carmo2020ptt5} or adapt and distill \cite{yao2021adapt} can be used. To pretrain a domain-specific model with limited domain-specific corpus, simultaneous pretraining which leverages both general and in-domain corpus can be used \cite{wada2020pre}. 

\textit{5. Choose the pretraining dynamics} – BERT model is pretrained on sentence pairs with static masking in small batch sizes.  Liu et al. \cite{liu2019roberta} showed that carefully designed pretraining choices like dynamic masking, large batch sizes, more pretraining steps, and long input sequences further enhance the performance of the model. Moreover, when using large batch sizes which may cause difficulty in optimization,  it is recommended to a)  linearly increase the learning rate in the early pretraining steps and b) use different learning rates in different layers which can also help to speed up convergence \cite{you2019large}.

\begin{figure*}[h]
\begin{center}
\includegraphics[width=18cm, height=8cm]{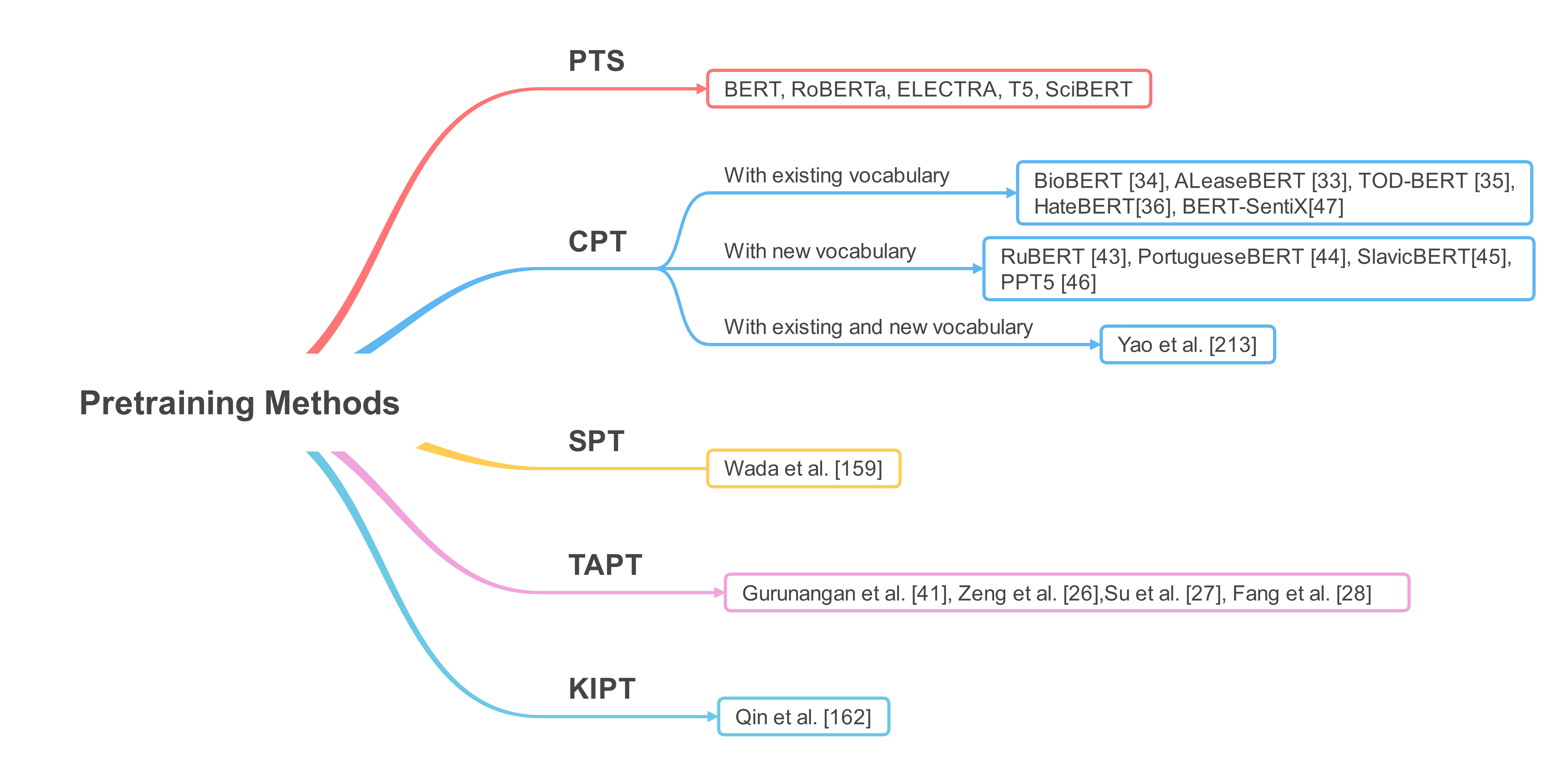}
\caption{\label{pretraining-methods-diag} Pretraining methods } 
\end{center}
\end{figure*}

 \subsubsection{Pretraining Corpus}
 
 \begin{table*}[h!]
\begin{center}
{\renewcommand{\arraystretch}{1.5}
\begin{tabular}{|p{2cm}|p{2cm}|p{2cm}|p{6.5cm}|p{3.5cm}|}
\hline
  \multicolumn{2}{|c|}{\textbf{Pretraining Corpus Type}} & \textbf{Pretraining Corpus} & \textbf{Description} & \textbf{Models}  \\
  \hline
 \multirow{2}{*}{General} & -  &	OpenWebText &	Open source equivalent to the WebText corpus used to pretrain GPT-2 model and it is around 32GB.  & RoBERTa \cite{liu2019roberta} \\ \cline{2-5} 
 
 & - & C4 \cite{raffel2019exploring} &	750GB collection of common crawl text which is deduplicated and filtered to include natural text only. &	T5 \cite{raffel2019exploring} \\ \hline
 
 \multirow{2}{*}{Social Media}	& -	& Rale-E \cite{caselli2020hatebert} &	Collection of hateful comments in English which are posted on Reddit, a popular social media platform. &	HateBERT \cite{caselli2020hatebert} \\ \cline{2-5} 
 
&	-	& Amazon reviews \cite{ni2019justifying}	& Collection of 233M reviews posted by users about various products. The gathered reviews cover around 29 domains. &	BERT-SentiX \cite{zhou2020sentix} \\ \hline

\multirow{3}{*}{Domain-Specific} &	Biomedical &	MIMIC-III \cite{johnson2016mimic} &	Consists of de-identified ICU patient records gathered over more than a decade. It is the largest publicly available corpus of clinical records. &	BioBERT \cite{lee2020biobert}, BlueBERT \cite{peng2019transfer}, ClinicalBERT \cite{alsentzer2019publicly} \\ \cline{2-5} 

& News &	RealNews \cite{zellers2019defending}	& Common crawl new corpus. It is around 120GB. &	Roberta-base-news \cite{gururangan2020don} \\ \cline{2-5} 

&	Academic &	S2ORC \cite{lo2020s2orc} &	Large scale collection of more than 80M research papers written in English.  &	Roberta-base-biomed \cite{gururangan2020don}, Roberta-base-cs \cite{gururangan2020don} \\ \hline

\end{tabular}}
\end{center}
\caption{ \label{taxonomy-corpora-1}  Summary of various pretraining corpora in general, social media and specific domains.} 
\end{table*}

 Self-Supervised learning to pretrain T-PTLMs requires large volumes of pretraining data. As shown in Figure, pretraining corpus can be classified into four types (refer Figure \ref{pretraining-corpora}). The characteristic of the text differs from one type of corpus to another. For example, in the general domain, the text is less noisy and written formally by professionals. In social media, the text is mostly noisy and written colloquially by the general public. Moreover, many specific domains like Biomedical and Finance contain many domain-specific words which are not used in the general domain. In general, the performance of general domain models in domain-specific tasks is limited \cite{lee2020biobert}. So, we have to choose the pretraining corpus depending on the target domain to achieve good results. BERT model is pretrained using text from Wikipedia and BookCorpus which amounts to 16GB \cite{devlin2019bert}. Further research works showed that the performance of the model can be increased by using large pretraining datasets \cite{liu2019roberta,yang2019xlnet}. This triggered the development of much larger datasets, especially from the common crawl. For example, C4 data contains around 750GB of text data \cite{raffel2019exploring} while CC-100 corpus includes around 2.5TB of text data \cite{conneau2020unsupervised}. Multilingual T-PTLMs like mBERT \cite{devlin2019bert}, IndT5 \cite{chen2021indt5}, IndoBART \cite{cahyawijaya2021indonlg}, and XLM-R \cite{conneau2020unsupervised} are pretrained using only multilingual datasets. Some of the models like XLM \cite{lample2019cross}, XLM-E \cite{chi2021xlm}, infoXLM \cite{chi2020infoxlm}, and mT6 \cite{chi2021mt6} are pretrained using both multilingual and parallel datasets. A summary of various pretraining corpora is given in Tables \ref{taxonomy-corpora-1} and \ref{taxonomy-corpora-2}.
 
 \begin{table*}[h!]
\begin{center}
{\renewcommand{\arraystretch}{1.5}
\begin{tabular}{|p{2cm}|p{2cm}|p{2cm}|p{6.5cm}|p{3.5cm}|}
\hline
\multicolumn{2}{|c|}{\textbf{Pretraining Corpus Type}} & \textbf{Pretraining Corpus} & \textbf{Description} & \textbf{Models}  \\ \hline
  
\multirow{2}{*}{Language-based}	& Monolingual (Indonesian) &	Indo4B \cite{wilie2020indonlu} &	Collection of 23GB of Indonesian text gathered from various public resources including social media platforms. It includes around 3.58 billion words. &	IndonesianBERT \cite{wilie2020indonlu} \\ \cline{2-5}

&	Monolingual (Portuguese) &	BrWaC \cite{wagner2018brwac} &	17.5GB collection of Brazilian Portuguese web text gathered from 3.5 million web pages. It includes around 2.7B words. &	PortugueseBERT \cite{souza2020bertimbau}, PTT5 \cite{carmo2020ptt5} \\ \cline{2-5}

&	Monolingual (Chinese) &	CLUECorpus2020 \cite{xu2020cluecorpus2020}	& Collection of 100GB of Chinese common crawl text. &	RoBERTa-tiny-clue \cite{xu2020cluecorpus2020} \\ \cline{2-5}

&	Monolingual (Chinese) &	WuDaCorpora \cite{yuan2021wudaocorpora} &	200GB collection of Chinese Web text. It includes around 72B Chinese characters.	& Chinese-Transformer-XL \cite{yuan2021wudaocorpora} \\ \cline{2-5}

&	\multirow{7}{*}{Multilingual} &	Indo4Bplus \cite{cahyawijaya2021indonlg}	& Includes text from Indo4B corpus  for Indonesian and from Wikipedia, CC-100 for Sundanese and Javanese language. &	IndoBART \cite{cahyawijaya2021indonlg} \\ \cline{3-5}

&	 &	OSCAR \cite{suarez2019asynchronous} &	Large scale collection of common crawl text for around 166 languages. &	MuRIL \cite{khanuja2021muril}
\\ \cline{3-5}

&	 &	IndicCorp \cite{kakwani2020inlpsuite}	& Collection of text from various sources for 12 major Indian languages. It includes around 8.9B words. &	IndicBERT \cite{kakwani2020inlpsuite} \\ \cline{3-5}

&	 &	mC4 \cite{xue2021mt5} &	Multi-lingual equivalent of C4. It includes common crawl text for 101 languages. &	mT5 \cite{xue2021mt5} \\ \cline{3-5}

&	 &	CC-Net \cite{wenzek2020ccnet} &	Large scale collection of common crawl text for more than 100 languages. &	mT6 \cite{chi2021mt6} \\ \cline{3-5}

&	 &	CC-100 \cite{conneau2020unsupervised}	& Large scale collection (2.5TB) of common crawl text of 100 languages. &	XLM-R \cite{conneau2020unsupervised}, infoXLM \cite{chi2020infoxlm}, XLM-E \cite{chi2021xlm} \\ \cline{3-5}

&	 &	IndCorpus \cite{chen2021indt5} &	Collection of text from Wikipedia and Bible for 11 languages including Indigenous languages. It is around 1.17GB and includes around 5.37M sentences.	& IndT5 \cite{chen2021indt5} \\ \cline{2-5}

&	\multirow{7}{*}{Parallel}	& PMINDIA \cite{haddow2020pmindia}	& Collection of parallel data gathered from Prime Minister of India website. The corpus includes 56K sentences for each language pair. &	MuRIL \cite{khanuja2021muril} \\ \cline{3-5}

&	 &	IIT Bombay \cite{kunchukuttan2018iit} &	Collection of 1.49M English-Hindi parallel sentences. &  	mT6 \cite{chi2021mt6}, XLM \cite{lample2019cross}, infoXLM \cite{chi2020infoxlm}, Unicoder \cite{huang2019unicoder}, ALM \cite{yang2020alternating}, XLM-E \cite{chi2021xlm} \\ \cline{3-5}

&		& Multi-UN \cite{ziemski2016united} &	Parallel corpus created from UN official documents for six languages. & 	mT6 \cite{chi2021mt6}, XLM \cite{lample2019cross}, infoXLM \cite{chi2020infoxlm}], Unicoder \cite{huang2019unicoder}, ALM \cite{yang2020alternating}, XNLG \cite{chi2020cross}, XLM-E \cite{chi2021xlm} \\ \cline{3-5}

&	 &	Wiki-Matrix \cite{schwenk2021wikimatrix} & 	Parallel data for 85 languages. It includes 135M parallel sentences in 1620 language pairs out of which 34M sentences are aligned with English. &	mT6 \cite{chi2021mt6}, XLM-E \cite{chi2021xlm} \\ \cline{3-5}

&	 &	CC-Aligned \cite{el2020massive}	& Parallel corpus of 292 million non-English common crawl document pairs and 100 million English common crawl document pairs. & 	XLM-E \cite{chi2021xlm} \\ \cline{3-5}

&	 &	Dakshina \cite{roark2020processing} &	Parallel corpus containing 10K sentences for 12 Indian languages. Each sentences consists of sentence in  native script sentence and its manually romanized transliteration. &	MuRIL \cite{khanuja2021muril} \\ \cline{3-5}

&	 &	Samanantar \cite{ramesh2021samanantar} &	Includes 49.6M sentences pairs of 12 Indian languages aligned with English. It is the largest publicly available parallel corpus for Indian languages.	& - \\ \hline

\end{tabular}}
\end{center}
\caption{\label{taxonomy-corpora-2}  Summary of various language-based pretraining corpora.} 
\end{table*}

\subsection{Types of Pretraining Methods}
Figure \ref{pretraining-methods-diag} shows the classification of pretraining methods into five types.
\label{pretraining-methods}

\subsubsection{Pretraining from Scratch (PTS)} 
\begin{figure}[t!]
\begin{center}
\includegraphics[width=5cm, height=1cm]{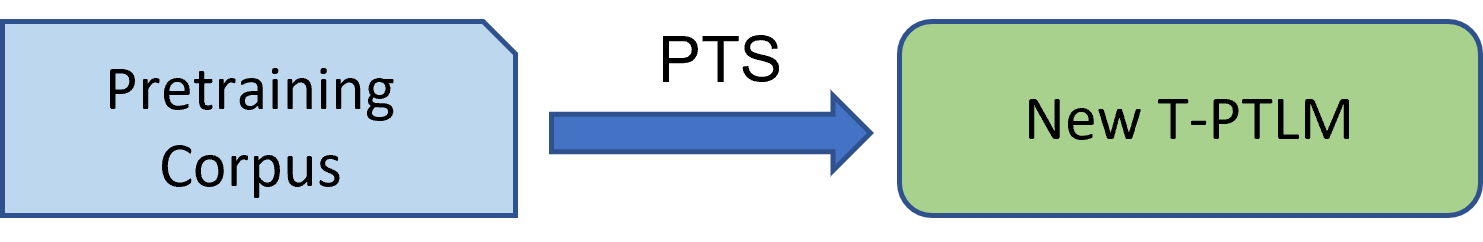}
\caption{\label{PTS} Pretraining from Scratch (PTS) } 
\end{center}
\end{figure}

Models like BERT, RoBERTa , ELECTRA, and T5  are pretrained from scratch on large volumes of unlabeled text (refer Figure \ref{PTS}). Usually, any transformer-based pretrained language model consists of an embedding layer, transformer encoder, or (and) transformer decoder layers. All these layer parameters are randomly initialized and then learned during pretraining by minimizing the losses of one or more pretraining tasks. For example, BERT model is pretrained from scratch using MLM and NSP. Pretraining from scratch is computationally expensive and requires a large number of GPUs or TPUs.

\subsubsection{Continual Pretraining (CPT)} 
\begin{figure}[t]
\begin{center}
\includegraphics[width=7cm, height=2cm]{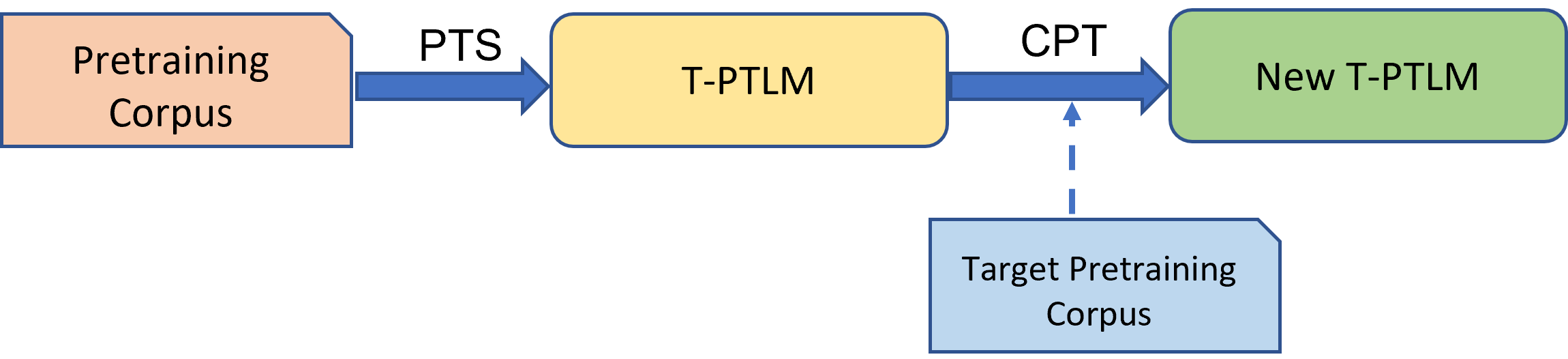}
\caption{\label{CPT} Continual Pretraining (CPT) } 
\end{center}
\end{figure}

Models like BioBERT \cite{lee2020biobert}, ALeaseBERT \cite{leivaditi2020benchmark}, TOD-BERT \cite{wu2020tod}, HateBERT \cite{caselli2020hatebert}, infoXLM \cite{chi2020infoxlm}, and XNLG \cite{chi2020cross} are obtained by initializing from existing pretrained models and then further pretrained. For example, infoXLM is initialized from XLM-R \cite{conneau2020unsupervised} and further pretrained on both monolingual and parallel data, ALeaseBERT is initialized from general ALBERT and further pretrained on lease agreements, XLM \cite{lample2019cross} parameters are used to initialize both encoder and decoder layers in XNLG. Unlike in PTS, in continual pretraining the model parameters are not learned from scratch. Instead, the parameters are initialized with existing language model parameters and then adapted to the target domain by further pretraining (refer Figure \ref{CPT}). Continual pretraining is commonly used to develop T-PTLMs in specific domains like social media \cite{caselli2020hatebert,barbieri2020tweeteval,barbieri2021xlm}, biomedical \cite{lee2020biobert}, Legal \cite{leivaditi2020benchmark}, News \cite{gururangan2020don}, Computer Networking \cite{louis2020netbert} etc. The main advantage of continual pretraining is that it avoids training a model from scratch and makes use of the existing language model parameters. As CPT starts from existing model parameters, it is less expensive and requires less training time and computational resources compared to PFS. 

However, the lack of target domain-specific vocabulary is a drawback in CPT when the target domain consists of many domain-specific words. For example, BioBERT \cite{lee2020biobert} is initialized from general BERT and further pretrained on biomedical text. Though the language model is adapted to the biomedical domain, the vocabulary which is learned over general domain text does not include many of the domain-specific words. As a result, domain-specific words are split into a number of sub-words which hinders model learning and degrades its performance in downstream tasks. Similarly, mBERT accommodates more than 100 languages, the number of tokens in its vocabulary (110K) specific to a language is less. A possible solution for this is continual pretraining with target domain or language-specific vocabulary \cite{souza2020bertimbau,arkhipov2019tuning,carmo2020ptt5}. Here, new vocabulary is generated over the target domain or language text. During continual pretraining, the embedding layer is randomly initialized and all other layer parameters are initialized with existing language model parameters. For example, models like RuBERT \cite{kuratov2019adaptation}, PortugueseBERT \cite{souza2020bertimbau}, SlavicBERT \cite{arkhipov2019tuning} are initialized from mBERT but further pretrained with language-specific vocabulary. Similarly, PPT5 \cite{carmo2020ptt5} is initialized from the T5 model but further pretrained with language-specific vocabulary. However, the performance of the model obtained by continual pretraining with new vocabulary is slightly less but on par with the performance of the model trained from scratch. As CPT is computationally less expensive, CPT with new vocabulary can be preferred over PTS in resource-constrained situations.  Recently, Yao et al. \cite{yao2021adapt} proposed Adapt and distill approach to adapt general models to a specific domain using vocabulary expansion and knowledge distillation. Different from existing adaptation methods, Adapt and distill approach not only adapt general models to specific domain but also reduces the size of the model.

It is not necessary to use the same set of pretraining tasks used by the existing model for continual pretraining. For example, BERT-SentiX \cite{zhou2020sentix} model is initialized from BERT and further pretrained on product reviews using four sentiment-aware pretraining tasks.  Similarly, TOD-BERT \cite{wu2020tod} is initialized from BERT and further pretrained on dialogue corpus text using MLM and response contrastive loss (RCL).

\subsubsection{Simultaneous Pretraining (SPT)} 
\begin{figure}[h]
\begin{center}
\includegraphics[width=7cm, height=1.2cm]{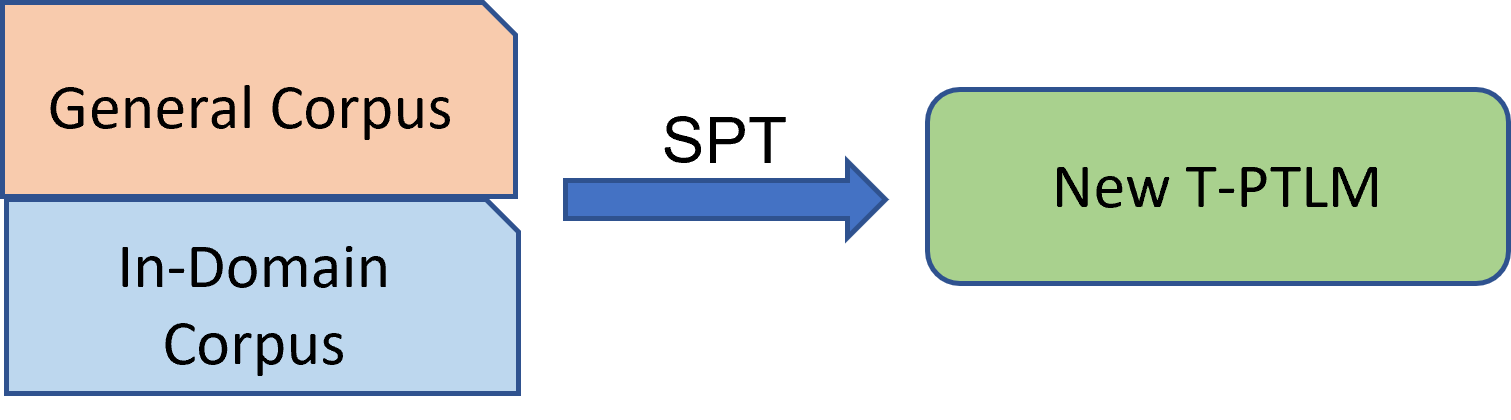}
\caption{\label{SPT} Simultaneous Pretraining (SPT) } 
\end{center}
\end{figure}

Domain-specific T-PTLMs can be developed by training from scratch or by continual pretraining. Both these pretraining methods require large volumes of domain-specific unlabelled text to pretrain the model. However, the availability of domain-specific text is limited in many domains. Moreover, domain-specific text in languages other than English is available in small quantities only. For example, in the biomedical domain, MIMIC-III \cite{johnson2016mimic} is the largest publicly available (English) medical records dataset. However, it is difficult to obtain such large volumes of medical records in languages like Japanese \cite{wada2020pre}.  PTS or CPT using a small amount of domain-specific text overfits the model. Simultaneous pretraining (SPT) allows the model to pretrain from scratch using a corpus having both general and domain-specific text \cite{wada2020pre} (refer Figure \ref{SPT}). Here, up sampling of domain-specific text is done to ensure a good number of domain-specific terms in model vocabulary and also to have a balanced pretraining. Wada et al. \cite{wada2020pre} showed that Japanese clinical BERT pretrained using SPT outperforms Japanese clinical BERT trained from scratch.  

\subsubsection{Task Adaptive Pretraining (TAPT)} 
\begin{figure}[h]
\begin{center}
\includegraphics[width=7cm, height=1.6cm]{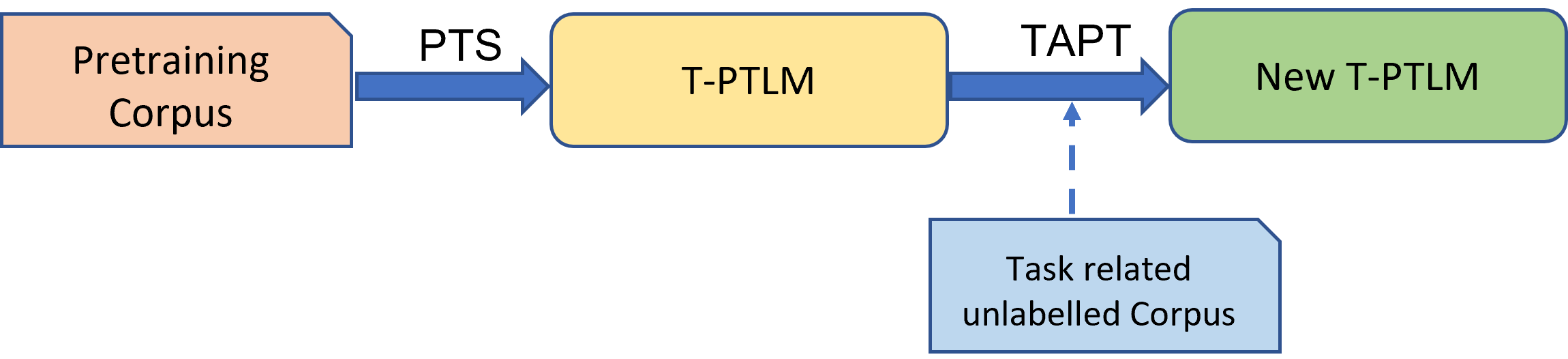}
\caption{\label{TAPT} Task Adaptive Pretraining (TAPT) } 
\end{center}
\end{figure}

Pretraining approaches like PTS, CPT and SPT allow the model to learn universal or domain-specific language representations by training on large volumes of general or domain-specific or combined text. As all these approaches involve training over a large amount of text, these approaches are expensive. Task Adaptive Pretraining (TAPT) allows the model to learn fine-grained task-specific knowledge along with domain-specific knowledge by pretraining on a small amount of task-specific unlabelled text \cite{gururangan2020don} (refer Figure \ref{TAPT}). As TAPT requires only a small amount of text, it is less expensive compared to other pretraining methods. Additional task-related sentences can be obtained from large domain corpus using lightweight approaches like VAMPIRE \cite{gururangan2019variational} which embeddings all the sentences using a simple bag-of-words language model.  Gururangan et al. \cite{gururangan2020don}  showed that TAPT is complementary to other pretraining approaches i.e., PTS / CPT followed by TAPT further improves the performance of the model. 

\subsubsection{Knowledge Inherited Pretraining (KIPT)}  
\begin{figure}[h]
\begin{center}
\includegraphics[width=7cm, height=1.6cm]{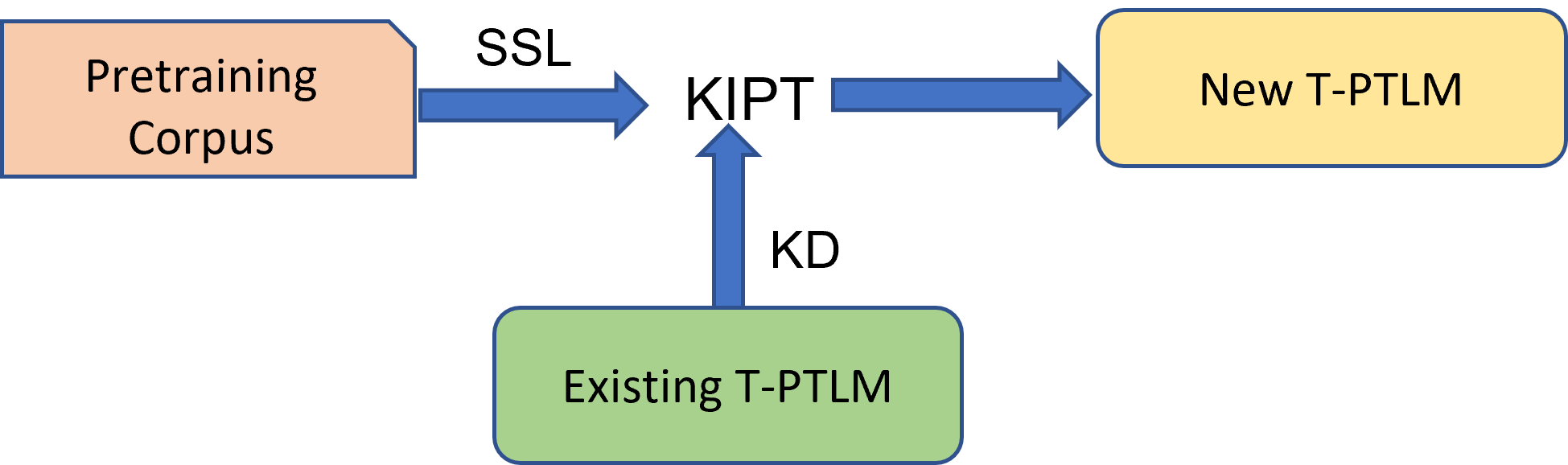}
\caption{\label{KIPT} Knowledge Inherited Pretraining (KIPT) } 
\end{center}
\end{figure}

All the previously discussed pretraining methods like PTS, CPT, SPT, and TAPT solely depend on self-supervised learning to pretrain the models. It is highly expensive and time-consuming to pretrain a large model from scratch using SSL only. In general, humans learn not only learn through self-learning but also learn from other knowledgeable people. Inspired from this,  Qin et al. \cite{qin2021knowledge} proposed Knowledge Inherited Pretraining (KIPT), a novel pretraining method which pretrains the model using both self-supervised learning and knowledge distillation (refer Figure \ref{KIPT}) . KIPT allows reusing the knowledge available in existing pretrained models to pretrain a new model.   

\begin{equation}
    L_{KIPT} = \sigma * L_{SSL} + (1-\sigma) * L_{KD} 
\end{equation}
 where $L_{KIPT}$ represents the overall loss of KIPT, $L_{SSL}$ and $L_{KD}$ represents losses of self-supervised learning and knowledge distillation. KIPT is similar to KD in reusing the knowledge from existing models. However, it is different from KD in two aspects, (a) in KD, generally the student model is compact in size compared to teacher model whereas in KIPT the student model is larger in size compared to teacher model (b) in KD, the student model solely learns from teacher model where as in KIPT the student model encodes the knowledge available in pretraining corpus using self-supervised learning in addition to the knowledge from teacher model. By learning from a knowledgeable teacher model along with self-supervised learning, the model learns more as well as converges faster which makes KIPT more effective and less expensive compared to the pretraining methods which involves only self-supervised learning. Due to the additional knowledge gained from a knowledgeable teacher model, the models trained using KIPT outperforms models trained using self-supervised learning only \cite{qin2021knowledge}. Further, Qin et al. \cite{qin2021knowledge} showed that KIPT supports both life-long learning and knowledge transfer. CPM-2 \cite{zhang2021cpm} is the first large scale pretrained language model pretrained using knowledge inheritance.

 \subsection{Pretraining Tasks}
 SSL allows T-PTLMs to learn universal language representations by solving one or more predefined tasks. These tasks are referred to as “pretext” or “pretraining” tasks. Pretraining tasks are self-supervised i.e., these tasks make use of pseudo labeled data. The data attributes and pretraining task definition determine the pseudo labels. A pretraining task should be challenging enough so that it provides more training signals to the model. For example, tasks like MLM involves only 15\% of tokens in each training sample for learning while tasks like Replaced Token Detection (RTD) \cite{clark2019electra}, Random Token Substitution (RTS) \cite{di2021efficient}, and Shuffled Token Detection (STD) \cite{panda2021shuffled} involves all the tokens in the input sample for model learning. Moreover, a pretraining task should be similar to the downstream task. For example, pretraining tasks like Seq2SeqLM \cite{song2019mass} or Denoising Auto Encoder (DAE) \cite{lewis2020bart} are similar to downstream tasks like text summarization, machine translation, etc. 
 
 \textbf{Casual Language Modeling (CLM)} - CLM or simply Unidirectional LM predicts the next word based on the context. The unidirectional LM can handle the sequence from left-to-right or right-to-left. In let-to-right LM, the context includes all the words on the left side while in right-to-left LM, the context includes all the words on the right side. GPT-1 \cite{radford2018improving} is the first transformer-based PTLM to use CLM (left-to-right) as a pretraining task. UniLM \cite{dong2019unified} uses both left-to-right and right-to-left CLM as pretraining tasks. Let $x=\{x_1,x_2,x_3,…,x_{|x|}\}$ represents a sequence where $|x|$ represents the number of tokens in the sequence. CLM loss is defined as 
 \begin{equation}
     L_{CLM}^{(x)}= -\frac{1}{|x|}  \sum_{i=1}^{|x|}logP( x_i/x_{<i})
 \end{equation}
 
 where $x_{<i}= x_1,x_2,x_3,..x_{i-1}$.
 
 \textbf{Masked Language Modeling (MLM)} – The main drawback in CLM is the inability to leverage both contexts. Bidirectional contextual information is much better compared to unidirectional context information for encoding token representations. It is not possible to train standard CLM using bidirectional context as it would allow a token to see itself which makes the prediction trivial. MLM is an improved version of CLM to leverage tokens from both contexts. In MLM,  we feed the masked token vectors to the softmax layer to get the probability distribution over the vocabulary and then use the cross-entropy loss. BERT is the first model to use MLM as pretraining task \cite{devlin2019bert}. The authors of BERT masked the tokens at a probability of 0.15. Let $x_{\setminus M_x }$ represents the masked version of $x$ and $M_x$ represents the set of masked token positions in $x$. MLM loss is defined as 
 \begin{equation}
     L_{MLM}^{(x)}= -\frac{1}{|M_x|} \sum_{i \in M_x}logP(x_i/x_{\setminus M_x}) 
 \end{equation}
 
 \textbf{Replaced Token Detection (RTD)} -   MLM is better than CLM by leveraging bidirectional contextual information. However, MLM has two drawbacks a) provides less training signal – in MLM, the model learns from only 15\% of the tokens and b) model see special mask token only during pretraining which results in a discrepancy between pretraining and fine-tuning stages. RTD overcomes these two issues by a novel approach that involves identifying the replaced tokens \cite{clark2019electra}. MLM corrupts the sentence by using special mask tokens while RTD corrupts the sentence using the output tokens from the generator model trained using MLM objective. MLM involves predicting the original tokens based on masked token vectors while RTD is a token-level binary classification task that involves classifying every token as replaced or not. ELECTRA model is pretrained in two steps. 1)train the generator model using MLM objective and 2)train the discriminator model initialized from a generator using RTD objective. Let $\hat{x}$ is the corrupted version of $x$. RTD loss is defined as
 \begin{equation}
     L_{RTD}^{(x)}= -\frac{1}{|\hat{x}|}  \sum_{i=1}^{|\hat{x}|} logP(d⁄\hat{x}_i)
\end{equation}
 where $d \in \{0,1\}$ represents whether the token is replaced or not. 
 
\textbf{Shuffled Token Detection (STD)} – STD is a token-level discriminative task that involves identifying the shuffled tokens. Similar to RTD, it is sample efficient and avoids discrepancy between pretraining and fine-tuning stages. In STD, the words are shuffled at a probability of 0.15 (this is based on the masking probability used in BERT and RoBERTa models). Panda et al. \cite{panda2021shuffled} showed that continual pretraining RoBERTa using STD improves its performance in many of the GLUE tasks and established that STD allows the model to learn more coherent sentence representations. Let $\hat{x}$ is the corrupted version of $x$. RTD loss is defined a

\begin{equation}
     L_{STD}^{(x)}= -\frac{1}{|\hat{x}|}  \sum_{i=1}^{|\hat{x}|} logP(d⁄\hat{x}_i)
\end{equation}
 where $d \in \{0,1\}$ represents whether the token is shuffled or not.
 
 \textbf{Random Token Substitution (RTS)} – RTD is sample efficient but requires a separate generator to corrupt the input sequence. Training a separate generator model is computationally expensive. To overcome this drawback, Di et al. \cite{di2021efficient} proposed RTS which involves identifying the randomly substituted tokens. In RTS, 15\% of the tokens are randomly substituted with other tokens from the vocabulary. RTS is sample efficient like RTD but does not require any separate generator model to corrupt the input sequence. Di et al.\cite{di2021efficient}  showed that RoBERTa model trained using RTS matches the performance of RoBERTa model trained using MLM while requiring less training time.  RTS loss is defined as 
  \begin{equation}
     L_{RTS}^{(x)}= -\frac{1}{|\hat{x}|}  \sum_{i=1}^{|\hat{x}|} logP(d⁄\hat{x}_i)
\end{equation}
 where $d \in \{0,1\}$ represents whether the token is randomly substituted or not and $\hat{x}$ is obtained by randomly substituting 15\% of tokens in $x$. 
 
 \textbf{Swapped Language Modeling (SLM)} – The MLM pretraining task uses a special mask token to corrupt the input sequence. However, the use of this special token results in a discrepancy between pretraining and fine-tuning stages. SLM overcomes this drawback by corrupting the sequence with random tokens from vocabulary at a probability of 0.15 \cite{di2021efficient}. SLM is similar to MLM by predicting the corrupting tokens but unlike MLM, SLM replaces the tokens with random tokens. SLM is similar to RTS in using the random tokens for corruption but unlike RTS which is sample efficient by involving every token in the input sequence, SLM is not sample efficient as it involves only 15\% of input tokens. SLM loss is defined as  
 
 \begin{equation}
     L_{SLM}^{(x)}= -\frac{1}{|R_x|} \sum_{i \in R_x}logP(x_i/x_{\setminus R_x}) 
 \end{equation}
Where $R_x$ represents set of positions of randomly substituted tokens and $x_{\setminus R_x}$ represents the corrupted version of x.

\textbf{Translation Language Modeling (TLM)} – TLM is an extension of MLM to use parallel data in cross-lingual pretraining. XLM \cite{lample2019cross} is the first cross-lingual model to use TLM as a pretraining task followed by XNLG \cite{chi2020cross}. TLM is also referred to as cross-lingual MLM (XMLM). Here the input is a pair of sentences ($x$,$y$) where $x$ and $y$ are parallel sentences i.e., $x$ is a translation of $y$. Similar to MLM, tokens from both sentences are randomly masked. As prediction of masked tokens involves context from both the sentences, TLM helps the model to learn cross-lingual mapping. TLM loss is similar to MLM and is defined as
\begin{multline}
     L_{MLM}^{(x,y)}= -\frac{1}{|M_x|} \sum_{i \in M_x}logP(x_i/x_{\setminus M_x},y_{\setminus M_y}) \\ -\frac{1}{|M_y|} \sum_{i \in M_y}logP(y_i/x_{\setminus M_x},y_{\setminus M_y})
\end{multline}
Where $M_x$ and $M_y$ represents the set of masked positions in the sentences x and y respectively, $x_{\setminus M_y}$  and $y_{\setminus M_y}$ represent the masked version of x and y respectively. 

\textbf{Alternate Language Modeling (ALM)}: ALM is a pretraining task to train cross-lingual language models. ALM involves predicting the masked tokens in the code-switched sentences generated from parallel sentences \cite{yang2020alternating}. For a given parallel sentence pair ($x$,$y$), a code-switched sentence is generated by randomly substituting some phrases of $x$ with their translations from $y$. ALM follows the same settings of standard MLM for masking the tokens. By pretraining the model on code-switched sentences, the model learns relationships between languages in a much better way. Yang et al. \cite{yang2020alternating} showed that cross-lingual model pretrained using ALM outperforms XLM which shows that ALM is a better alternative to TLM for pretraining cross-lingual language models. ALM loss is  defined as
\begin{equation}
    L_{ALM}^{(z(x,y))}= -\frac{1}{|M|}  \sum_{i \in M}logP(z_i /Z_{\setminus M}) 
\end{equation}
Where $z$ is the code-switched sentence generated from $x$ and $y$, $z_{\setminus M}$ represents the masked version of $z$ and $M$ represents the set of masked token positions in $z_{\setminus M}$.

\textbf{Sentence Boundary Objective (SBO)} – SBO pretraining task involves predicting the masked tokens based on the span boundary tokens and position embeddings \cite{joshi2020spanbert}. SBO is similar to MLM in predicting the masked tokens. However, it is different from MLM in three aspects (a) SBO masks only contiguous span of tokens while MLM masks tokens randomly and (b) in SBO, the prediction of masked tokens involves span boundary tokens and position embeddings while in MLM, the prediction of masked tokens involves only the masked token vectors. Moreover, SBO is much challenging compared to standard MLM as it is difficult to predict the entire span “an American football game” compared to predicting “game” when “an American football” is already known \cite{joshi2020spanbert}. SBO helps the model to perform better in downstream tasks like entity extraction, coreference resolution, and question answering which involves span-based extraction. SBO loss is defined as
\begin{equation}
    L_{SBO}^{(x))}= -\frac{1}{|S|}  \sum_{i \in S}logP(x_i /y_i) 
\end{equation}
where $y_i=f(x_{s-1},x_{e+1},p_{s-e+1})$ and $f$() is a two-layered feedforward neural network, S represents the positions of tokens in contiguous span, $|S|$ represents the length of span, $s$ and $e$ represent the start and end positions of span, $p$ represents the position embedding. 

\begin{figure*}[h]
\begin{center}
\includegraphics[width=18cm, height=8cm]{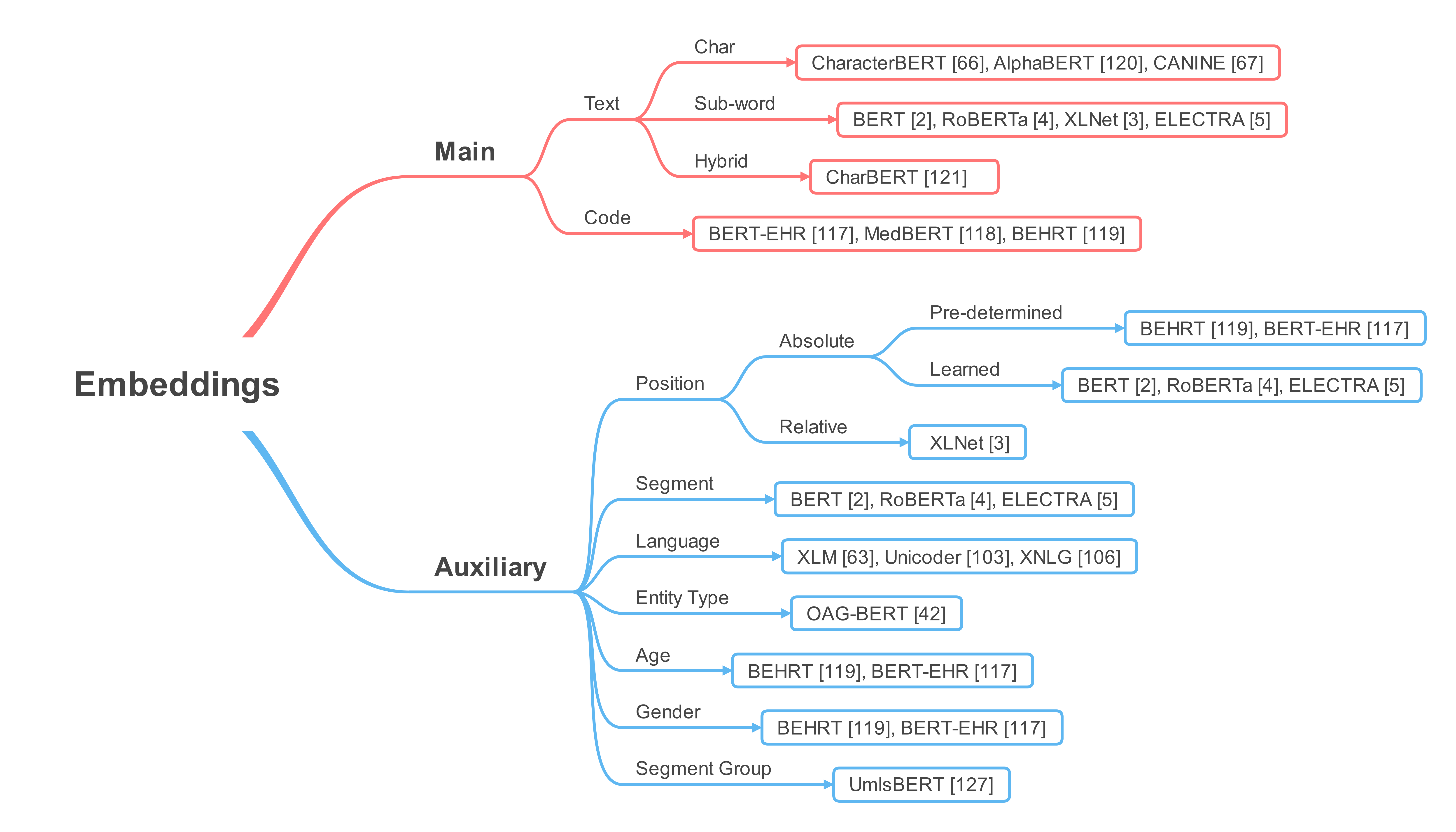}
\caption{\label{embeddings} Embeddings in T-PTLMs } 
\end{center}
\end{figure*}

\textbf{Next Sentence Prediction (NSP)} – NSP is a sentence-level pretraining task that helps the model to learn relationships between sentences \cite{devlin2019bert}. It is a binary sentence pair classification task that involves identifying consecutive sentences. Here the aggregate representation of the two sentences (x,y) i.e., [CLS] token vector is given to the sigmoid layer to get the probability.  For training, the sentence pairs are generated in a way that 50\% of instances are consecutive and the rest are not consecutive. Pretraining the model at the sentence level is useful in downstream tasks like question answering, NLI, and STS which involve sentence pair input. NSP loss is defined as 
\begin{equation}
     L_{NSP}^{(x,y)}= - logP(d⁄x,y)
\end{equation}
Where $d \in \{1,0\}$ represents whether the sentences are consecutive or not. 

\textbf{Sentence Order Prediction (SOP)} – NSP allows the model to learn sentence-level semantics and involves both topic and coherence prediction. As topic prediction is easier, the effectiveness of NSP is questioned \cite{liu2019roberta,lan2019albert}. SOP is a sentence-level pretraining task based on sentence coherence only. ALBERT \cite{lan2019albert} is the first pretraining model to use SOP as a pretraining task. It involves identifying whether the given sentences are swapped or not. Following NSP, the training instances are generated in a way that 50\% of instances are swapped and the rest or not. SOP loss is defined as
\begin{equation}
     L_{SOP}^{(x,y)}= - logP(d⁄x,y)
\end{equation}
Where $d \in \{1,0\}$ represents whether the sentences are swapped or not.

\textbf{Sequence-to-Sequence LM (Seq2SeqLM)} -  MLM is approached as a token level classification task over the masked tokens i.e., original words are predicted by feeding the masked token vectors to a softmax layer over the vocabulary. Seq2SeqLM is an extension of standard MLM to pretrain encoder-decoder-based models like T5 \cite{raffel2019exploring}, mT5 \cite{xue2021mt5} and MASS \cite{song2019mass}. In the case of MLM, the context includes all the tokens in the input sequence whereas in Seq2SeqLM, the context includes all the words in the input masked sequence and the left side words in the predicted target sequence. With masked sequence as input to the encoder, the decoder predicts the masked words from left to right sequentially. Seq2SeqLM loss is defined as
\begin{equation}
    L_{Seq2SeqLM}^{(x)}= -\frac{1}{l_s}  \sum_{s=i}^{j}logP(x_s⁄\hat{x},x_{i:s-1} )
\end{equation}
where $\hat{x}$  is the masked version of $x$ i.e., $x$ with masked n-gram span, $l_s$ represents the length of masked n-gram span.

\textbf{Denoising Auto Encoder (DAE)}: DAE helps to model to learn by reconstructing the original text from corrupted text \cite{lewis2020bart}. The text can be corrupted at token (e.g., token deletion and token masking), phrase (e.g., token infilling), sentence (e.g., sentence permutation), or document level (e.g., document rotation). Like Seq2SeqLM, DAE is useful to train encoder-decoder-based models. However, DAE is more sample efficient by providing more training signals for model learning. DAE provides more training signal as it involves reconstructing entire original text while Seq2SeqLM involves  reconstructing the masked tokens only. BART \cite{lewis2020bart} uses a bidirectional encoder to encode corrupted input sequence and a left-to-right decoder to recreate the original text. The authors of BART experimented with various corruption strategies and finally trained the model on the sentences corrupted using sentence permutation and text infilling. DAE loss is defined as
\begin{equation}
    L_{DAE}^{(x)}= -\frac{1}{|x|}   \sum_{i=1}^{|x|}logP(x_i⁄\hat{x},x_{<i})
\end{equation}
where $\hat{x}$ is the corrupted version of $x$.

\subsection{Embeddings}

Deep learning models expect the input in the form of a matrix of numbers and then apply a sequence of matrix operations. As deep learning models including transformers expect numerical input, input text should be mapped to a sequence of dense, low dimensional vectors (commonly called embeddings in natural language processing). In transformer-based pretrained language models, character or sub-word embeddings are preferred over word embeddings. This is because a) small vocabulary size in character and sub-word embeddings compared to word embeddings. The vocabulary of word embeddings consists of all the unique words (or all the words above the cut-off frequency) in the pretraining corpus, whereas vocabulary in character embedding models consists of all the characters and vocabulary in sub-word embedding models consists of all the characters, frequently occurring sub-words and words. The size of vocabulary also determines the overall size of pretrained language model \cite{ganesh2020compressing}. b) can represent any word and hence overcome the problem of OOV words which is a serious problem with word embeddings c) can encode fine-grained information at character or sub-word levels in word representation. Apart from representing input data using embeddings, it is also necessary additional information like position, language, etc. We classify embeddings into main and auxiliary depending on whether they represent the input data or provide additional information to the model (refer Figure \ref{embeddings}). 

\subsubsection{Main Embeddings} -  Main embeddings represent the input data in dense low dimensional vectors. In TPLMs, the input data is mostly a sequence of words. However, in domain-specific models like BERT-EHR \cite{meng2021bidirectional}, MedBERT \cite{rasmy2021med}, and BEHRT \cite{li2020behrt}   the input is a sequence of medical codes. All these three models are pretrained on medical text from electronic health records (EHRs).

\textbf{Text Embeddings}: Input for most of the TPLMs is the sequence of words. Input text can be represented using character, sub-word, or combination of character and sub-word embeddings.  Models like CharacterBERT \cite{el2020characterbert}, AlphaBERT \cite{chen2020modified} use character embeddings, and models like CharBERT \cite{ma2020charbert} use both character and sub-word embeddings. Models like BERT, RoBERTa, XLNet, T5, and BART use sub-word embeddings but the tokenizer used to generate the vocabulary is different in these models.

\textbf{Character Embeddings}: Character embeddings map each character to a dense low dimensional vector. The vocabulary of character embeddings includes all the characters like letters, symbols, punctuations, and numbers. Once the vocabulary is finalized, the embedding of each character in the vocabulary is randomly initialized and then learned during model pretraining. TPLMs use character embeddings in two ways. The first one is, character level word representation is generated from character embeddings and then a sequence of transformer layers are applied to encode contextual information \cite{el2020characterbert}. For example in CharacterBERT \cite{el2020characterbert}, fine-grained word representation is generated from character embeddings using a character encoder based on CharCNN and highway layer \cite{peters2018deep}. The second one is based on context string embedding \cite{akbik2018contextual}. Here, there is no notion of the explicit word, and input text is modeled as a sequence of characters. In AlphaBERT \cite{chen2020modified}, transformer layers are directly on character embeddings whereas in CharBERT \cite{ma2020charbert} transformer layers are applied after applying BiGRU on character embeddings. Here BiGRU processes the input at the character level and generates contextualized character embeddings \cite{cho2014learning}. 

\textbf{Sub-Word Embeddings}: Unlike character embeddings, the vocabulary of sub-word embeddings consists of characters, frequently occurring sub-words, and words. Here the vocabulary can be generated using any of the tokenizers like WordPiece \cite{wu2016google}, Byte Pair Encoding (BPE) \cite{sennrich2016neural}, Byte Level BPE (bBPE) \cite{radford2019language}, Unigram \cite{kudo2018subword}, and SentencePiece \cite{kudo2018sentencepiece}. Except Unigram, tokenizers like WordPiece, BPE, bBPE generate vocabulary by starting with base vocabulary having only the characters and iteratively augment the vocabulary until the predefined size is reached. BPE chooses the new symbol pair to be included in the vocabulary based on frequently while WordPiece does it based on language model probability. bBPE is the same as BPE except that it represents each character as a byte. Unigram starts with a large vocabulary and then arrives at a vocabulary of predefined size by iteratively cutting the characters which is exactly opposite to what happens in BPE and WordPiece.  Tokenizers like WordPiece and BPE assume space as a word separator in the input text which is not true in all cases. To overcome this, SentencePiece tokenizer treats space as a character and then generates the vocabulary using BPE or Unigram. 

The size of the vocabulary must be chosen carefully. Too small vocabulary size results in longer input sequences as more words will be split into many sub-words which hinders model learning and increases pretraining time. Too large vocabulary represents more words using a single token but increases the size overall of the model \cite{shin2020bio}. However, in the case of multilingual models like mBERT, XLM, and XLM-R, it is necessary to have a large vocabulary to accommodate more languages. Once the vocabulary is generated, each token in the vocabulary is assigned with a randomly initialized embedding and then learned during model pretraining.

\textbf{Hybrid Embeddings}: To leverage the benefits in both character and sub-word embeddings, models like CharBERT \cite{ma2020charbert} uses both character and sub-word embeddings. The model uses dual-channel CNN-based interaction module to model the interaction between character and sub-word embeddings. 

\textbf{Code Embeddings}: In the medical domain, each concept is represented using a standard code from ontology. Here the concept can be a disease, drug, symptom, etc. All the information during patient visits to a hospital is represented using medical codes in EHRs. Pretrained language models in the biomedical domain like BERT-EHR \cite{meng2021bidirectional}, MedBERT \cite{rasmy2021med}, and BEHRT \cite{li2020behrt} expect a sequence of medical codes as input. So, in these models vocabulary consists of medical codes from standard clinical ontologies. Embeddings for these medical codes are randomly initialized and learned during model pretraining.   

\subsubsection{Auxiliary Embeddings} Auxiliary embeddings provide additional information to the model. Each auxiliary embeddings have its purpose. For example, positional embeddings represent the position, while segment embeddings distinguish tokens from different sentences in the input sentence pair, language embeddings in multilingual pretrained models like XLM \cite{lample2019cross} and Unicoder \cite{huang2019unicoder} provide information about the language of the input sentence. Among auxiliary embeddings, position and segment embeddings are commonly used while the other embeddings are used in specific pretrained language models. For example, age, gender, and semantic group embeddings are used in biomedical pretrained language models only \cite{meng2021bidirectional,li2020behrt,michalopoulos2020umlsbert}. 

\textbf{Position Embeddings}: In traditional deep learning models like CNN and RNN, it is not necessary to provide any auxiliary embeddings along with text embeddings to represent the position of input tokens. This is because these models implicitly learn the order of input tokens. For example, RNN process input sequence character by character or word by word, and hence it automatically learns the order. In CNN, convolution operations are performed at a fixed size window level instead of character or word level and hence it also learns the order automatically. As transformers do not have convolution or recurrent layers to learn order, position embeddings are provided. Position embeddings can be absolute \cite{devlin2019bert,liu2019roberta,clark2019electra,meng2021bidirectional} or relative \cite{yang2019xlnet}. In models like BERT, RoBERTa, ELECTRA absolute position embeddings are learned along with other parameters of the model. However, in models like BERT-EHR \cite{meng2021bidirectional} and BEHRT \cite{li2020behrt}, absolute position embeddings are predetermined to handle an imbalance in patient sequence length.  

\textbf{Segment Embeddings}: In the case of sentence pair tasks, the model takes both the input sentences at the same time. So, it is necessary to distinguish tokens of two input sentences using segment embeddings. Position embedding is different for different tokens in the input sentence, but segment embedding is the same for all the tokens in each input sentence.  

\textbf{Language Embeddings}: Language embeddings are used in cross-lingual pretrained language models like XLM \cite{lample2019cross}, Unicoder \cite{huang2019unicoder}, and XNLG \cite{chi2020cross}. For example, models like XLM are pretrained using a) MLM on monolingual text data in 100 languages and b) TLM using parallel data. Language embeddings are used to explicitly inform the model about the language of the input sentence. In MLM which involves sentences in one language, language embedding will be the same for all the tokens in the input sentence. In the case of TLM which involves a pair of sentences from two different languages, language embedding will be the same for all the tokens in a sentence but different from the language embedding assigned to tokens in other input sentence. However, language embeddings are not used in XLM-R \cite{conneau2020unsupervised} model to allow the model to better deal with code-switching.

\textbf{Entity Type Embeddings}: OAG-BERT \cite{liu2021oag} is a pretrained academic language model like SciBERT \cite{beltagy2019scibert}. It is pretrained on academic text corpus. Unlike SciBERT which is just pretrained on academic text, OAG-BERT is pretrained on academic text as well as augmented with information about various entities in academic text like paper, published venue, author affiliation, research domain, and authors. Information about various entities is provided to the model during pretraining via entity type embeddings.   

\textbf{Age and Gender Embeddings}: In medical pretrained models like BEHRT \cite{li2020behrt}  and BERT-EHR \cite{meng2021bidirectional}, the input is a sequence of patient visits where each patient visit is represented as a sequence of medical codes. Apart from model codes, it is useful to provide additional information like age and gender. For example, each patient visits happen at different times. Providing age information to the model allows it to leverage temporal information. Age and gender information is provided to these models explicitly via age and gender embeddings. 

\textbf{Semantic Group Embeddings}: UmlsBERT \cite{michalopoulos2020umlsbert} is a knowledge enriched medical language model. It is obtained by continual pretraining ClinicalBERT \cite{alsentzer2019publicly} on UMLS data using novel multi-label MLM pretraining task. UMLS is a collection of over 100 medical ontologies. In UMLS, each medical concept is assigned a unique called Concept Unique Identifier, semantic type, synonyms, related concepts, etc. During continual pretraining, semantic type information is provided explicitly via semantic group embedding so that the language model can a) learn better representation for rare words and b) better model the association between words of the same semantic type.

 \section{Taxonomy}
 \label{taxonomy-sec}
\begin{figure*}[p]
\begin{center}
\includegraphics[width=19cm, height=22cm]{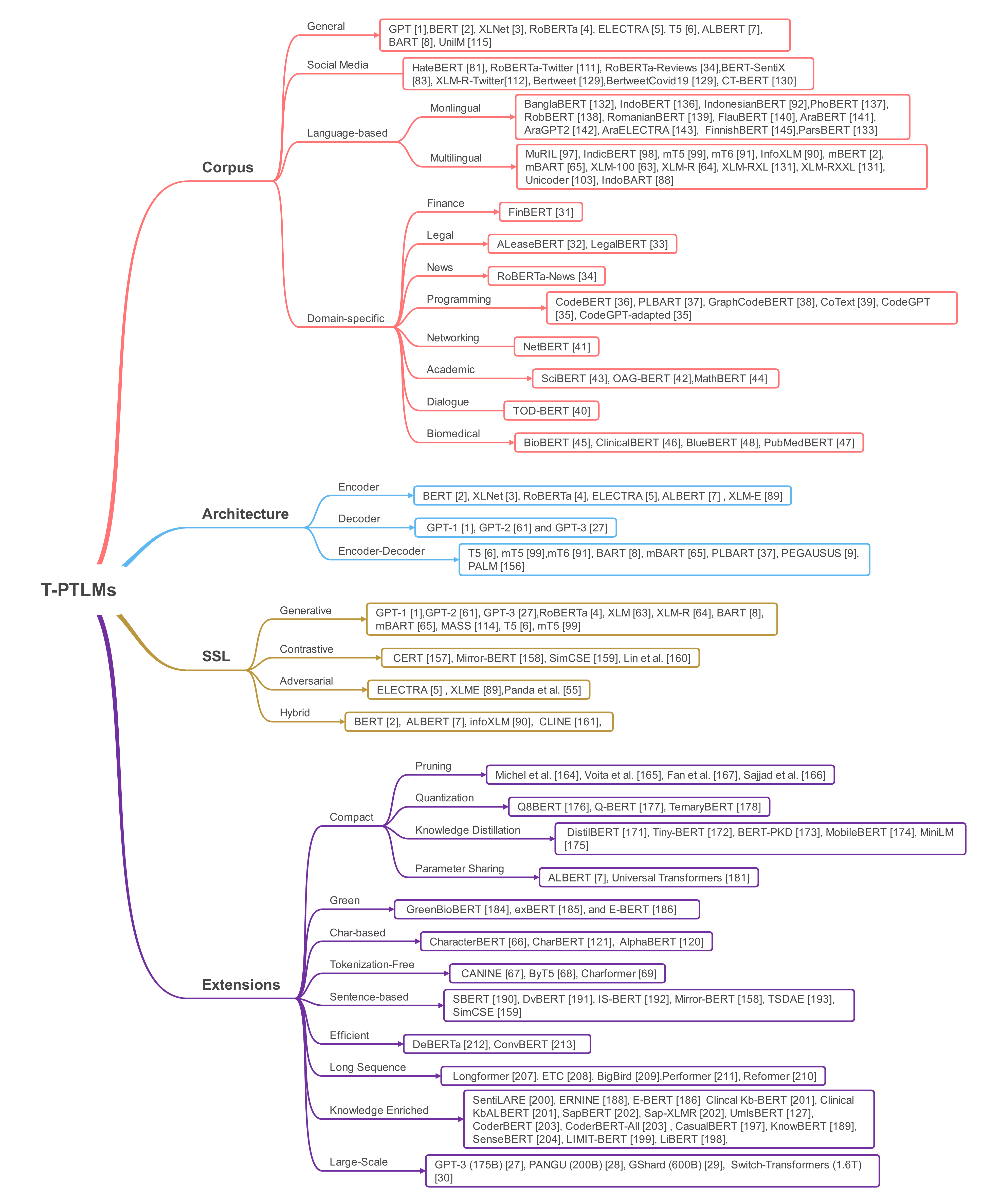}
\caption{\label{taxonomy} Taxonomy of T-PTLMs} 
\end{center}
\end{figure*}

 To understand and keep track of the development of various T-PTLMs, as shown in Figure \ref{taxonomy}, we classify T-PTLMs from four different perspectives namely Pretraining Corpus (Section \ref{corpus-based-sec}), Model Architecture (Section \ref{architecture-sec}), Type of SSL (Section \ref{type-ssl-sec}), and Extensions (Section \ref{extensions-sec}). 

\subsection{Pretraining Corpus-based}
\label{corpus-based-sec}
\subsubsection{General}
Models like GPT-1 \cite{radford2018improving}, BERT \cite{devlin2019bert}, UnilM \cite{dong2019unified}, XLNet \cite{yang2019xlnet}, RoBERTa \cite{liu2019roberta}, ELECTRA \cite{clark2019electra}, T5 \cite{raffel2019exploring}, and BART \cite{lewis2020bart} are pretrained on general corpus. For example, GPT-1 is pretrained on Books corpus while BERT and UniLM are pretrained on English Wikipedia and Books corpus. As the amount of text data available in Book corpus or English Wikipedia, text is gathered from multiple sources for pretraining models like XLNet, RoBERTa, ELECTRA, BART and T5. 

 \subsubsection{Social Media-based}

\begin{table*}[t!]
\begin{center}
{\renewcommand{\arraystretch}{1.5}
\begin{tabular}{|p{2.5cm}|p{2cm}|p{2.5cm}|p{5cm}|p{3.5cm}|}
\hline
  \textbf{Name} & \textbf{Pretrained from} & \textbf{Pretraining tasks} & \textbf{Corpus} & \textbf{Evaluation}  \\
  \hline
 HateBERT \cite{caselli2020hatebert} &	BERT &	MLM	& RAL-E (dataset of 1.5M hateful Reddit comments) &	Offensive tweets classification \\ \hline
 
RoBERTa-Twitter \cite{barbieri2020tweeteval} &	RoBERTa	& MLM &	Tweets (60M) &	Tweet classification \\ \hline

RoBERTa-Reviews \cite{gururangan2020don}	& RoBERTa &	MLM &	Amazon reviews (24.75M) \cite{he2016ups} &	Review classification \\ \hline

BERT-SentiX \cite{zhou2020sentix}	& BERT &	SWP, WP, EP and RP &	Amazon (233M) \cite{ni2019justifying} and Yelp reviews (8M) Reviews &	Cross domain sentiment analysis \\ \hline

XLM-R-Twitter \cite{barbieri2021xlm} &	XLM-R &	MLM	 & Tweets in multiple languages (198M) &	TweetEval \cite{barbieri2020tweeteval} and UMSAB \cite{barbieri2021xlm}  \\ \hline

Bertweet \cite{nguyen2020bertweet} &	Scratch & MLM	& Tweets (845M English + 5M COVID tweets) &	POS, NER and Tweets classification \\ \hline

BertweetCovid19 \cite{nguyen2020bertweet} &	Bertweet &	MLM &	COVID tweets (23M) &	Tweet classification \\ \hline

CT-BERT \cite{muller2020covid} &	BERT &	MLM, NSP &	COVID tweets (160M) &  Tweet classification \\ 
\hline
  
\end{tabular} 
}
\end{center}
\caption{\label{taxonomy-social}  Summary of social-media based T-PTLMs.} 
\end{table*}

T-PTLMs like BERT and RoBERTa are pretrained on formal text. As social media text is highly informal in nature with a lot of noise in the form of irregular grammar, slang words, and non-standard abbreviations, these models have limited performance on social media datasets \cite{caselli2020hatebert,nguyen2020bertweet,muller2020covid,barbieri2020tweeteval} . Researchers working at the intersection of social media and NLP have developed social media-specific T-PTLMs either by training from scratch \cite{nguyen2020bertweet} or continual pretraining \cite{caselli2020hatebert,muller2020covid,barbieri2020tweeteval,barbieri2021xlm,gururangan2020don,zhou2020sentix}  and a summary of these models is presented in Table \ref{taxonomy-social}.  Except for Bertweet \cite{nguyen2020bertweet}, all other social media-based T-PTLMs are developed by continual pretraining. Training from scratch is effective only when the pretraining corpus consists of a large number of tweets. Otherwise, continual pretraining is recommended. For example, BERTweet \cite{nguyen2020bertweet} is pretrained from scratch using 850M tweets. Barbieri et al. \cite{barbieri2020tweeteval} showed that RoBERTa model trained from scratch on tweets achieved less performance compared to RoBERTa model adapted to social media by continual pretraining. This is because of using just 60M tweets for pretraining. Different from other social media-based T-PTLMs which are developed using commonly used pretraining tasks like MLM and NSP, BERT-SentiX \cite{zhou2020sentix} is obtained by continual pretraining on user reviews using four novel sentiment aware pretraining tasks. 
 \subsubsection{Language-based}
 Language-based T-PTLMs can be monolingual or multi-lingual. Monolingual T-PTLMs are pretrained on specific language corpus while multi-lingual T-PTLMs are pretrained multiple language corpus.
 
 \textbf{Multi-lingual T-PTLMs} 
 Inspired by the tremendous success of BERT in English, the authors of BERT developed mBERT by pretraining BERT model from scratch using Wikipedia text from 104 languages \cite{devlin2019bert}. mBERT is the first multilingual T-PTLM. Following mBERT, many multilingual T-PTLMs are proposed. Recent research works showed that the performance of the model can be improved by training on large volumes of text. So, XLM-R \cite{conneau2020unsupervised} is pretrained on CC-100 which consists of a large amount of text particularly for low-resource languages compared to Wikipedia. Inspired from “Scaling laws for neural language models” \cite{kaplan2020scaling} much larger models like XLM-RXL \cite{goyal2021larger} and XLM-RXXL \cite{goyal2021larger} are pretrained on CC-100 and achieved much better results. mBERT, XLM-R, and its variants are pretrained on only non-parallel data. However, pretraining the model on parallel data along with non-parallel data allows the model to learn cross-lingual representations in a much better way. Models like MuRIL \cite{khanuja2021muril}, mT6 \cite{chi2021mt6}, InfoXLM \cite{chi2020infoxlm}, XLM \cite{lample2019cross} and Unicoder \cite{huang2019unicoder} are pretrained on both parallel and non-parallel data. Multi-lingual NLP research community also developed many generative T-PTLMs like mT5 \cite{xue2021mt5}, mT6 \cite{chi2021mt6}, mBART \cite{liu2020multilingual} and IndoBART \cite{cahyawijaya2021indonlg} based on encoder-decoder architecture. A summary of various multi-lingual T-PTLMs is presented in Table \ref{taxonomy-multi}. 
 
 \begin{table*}[t!]
\begin{center}
{\renewcommand{\arraystretch}{1.5}
\begin{tabular}{|p{2cm}|p{1.5cm}|p{3.2cm}|p{2.5cm}|p{2cm}|p{1cm}|p{2.5cm}|}
\hline
  \textbf{Name} & \textbf{Architecture} & \textbf{Pretraining tasks} & \textbf{Corpus} & \textbf{Vocabulary} & \textbf{\#Lang} & \textbf{\#Parameters}  \\
  \hline
 MuRIL \cite{khanuja2021muril} &	Encoder &	MLM + TLM &	Wikipedia,  Common Crawl + Parallel data &	WordPiece (197K) &	17	& 236M \\ \hline
 
IndicBERT \cite{kakwani2020inlpsuite} &	Encoder &	MLM	& IndicCorp &	SentencePiece (200K) &	12 &	33M \\ \hline

mT5 \cite{xue2021mt5} &	Encoder-Decoder	& Seq2SeqLM	& mC4 & 	SentencePiece (250K) &	101	& 300M, 580M, 1.2B, 3.7B and 13B (base, large, xl and xxl) \\ \hline

mT6 \cite{chi2021mt6} &	Encoder-Decoder &	Seq2SeqLM, MT, TPSC, TSC &	CCNet + Parallel data &	SentencePiece (250K) &	94 &	300M \\ \hline

InfoXLM \cite{chi2020infoxlm} &	Encoder &	MLM, TLM , XLCo &	CC-100 + Parallel data &	SentencePiece (250K) &
	94 &	270M and 559M (base and large) \\ \hline
	
mBERT \cite{devlin2019bert} &	Encoder	& MLM , NSP	& Wikipedia	& WordPiece (110K) &	104 &	172M \\ \hline

mBART \cite{liu2020multilingual} &	Encoder-Decoder &	DAE	& CC-25 &	SentencePiece (250K) &	25 &	680M \\ \hline

XLM-15 \cite{lample2019cross} &	Encoder &	MLM, TLM &	Wikipedia + Parallel data &	BPE (95K) &	15 &	250M \\ \hline

XLM-17 \cite{lample2019cross}	& Encoder &	MLM	& Wikipedia & 	BPE (200K) &	17	& 570M \\ \hline

XLM-100 \cite{lample2019cross} &	Encoder &	MLM &	Wikipedia & 	BPE (200K) &	100 &	570M \\ \hline

XLM-R \cite{conneau2020unsupervised} &	Encoder &	MLM	 & CC-100 &	SentencePiece (250K) &	100	& 270M ,560M (base and large) \\ \hline

XLM-RXL \cite{goyal2021larger} &	Encoder &	MLM &	CC-100 &	SentencePiece (250K) &	100 &	3.5B \\ \hline

XLM-RXXL \cite{goyal2021larger} &	Encoder & 	MLM	& CC-100 &	SentencePiece (250K) &	100 &	10.7B \\ \hline

Unicoder \cite{huang2019unicoder} &	Encoder & 	MLM, TLM, CLWR, CLPC, CLMLM	& Wikipedia + Parallel data &	BPE (95K) &	15 &	250M \\ \hline

IndoBART \cite{cahyawijaya2021indonlg} &	Encoder-Decoder &	DAE &	Indo4B-plus &	BPE (40K) &	3 &	130M \\ \hline

\end{tabular} }
\end{center}
\caption{\label{taxonomy-multi}  Summary of multi-lingual T-PTLMs.} 
\end{table*}
 
\textbf{Monolingual T-PTLMs}
 Multilingual models are pretrained on the corpus from multiple languages and hence they can be used for NLP tasks in more than one language. However, the following drawbacks force the NLP community to develop separate models for each language starting from BanglaBERT \cite{bhattacharjee2021banglabert} to ParsBERT \cite{farahani2020parsbert}.
 \begin{itemize}
     \item \textbf{Curse of Multilinguality} \cite{conneau2020unsupervised}: Multilingual models cannot represent all the languages equally. This is because of the underrepresentation of low-resource languages in the pretraining corpus and the limited capacity of the model. Moreover, adding more languages after a certain limit reduces the model performance.
     \item \textbf{Embedding barrier} \cite{bhattacharjee2021banglabert}: The performance of multilingual models in high resource languages that have adequate representation in model vocabulary is on par with their monolingual models \cite{rust2020good}. However, in the case of languages without adequate representation in model vocabulary, the difference in the performance of the monolingual and multilingual model is significant. Due to the high imbalance in the pretraining corpus, the representation of low-resource languages in multilingual model vocabulary is very limited which is referred to as the embedding barrier \cite{chi2020infoxlm}. For example, the representation of the Arabic language \cite{abdul2020arbert} in popular multilingual models is 5K out of 110K in mBERT and 14K out of 250K in XLM. This issue is more severe in the case of languages like Bangla that does not share vocabulary or script with any high-resource languages. The percentage of Bangla vocabulary in the multilingual model is less than 1\% \cite{bhattacharjee2021banglabert}. With very limited representation in the vocabulary, words in low resource languages are tokenized into many subwords which increases input sequence length, hinders model learning, and makes training expensive.  
 \end{itemize}

\begin{table*}[t!]
\begin{center}
{\renewcommand{\arraystretch}{1.5}
\begin{tabular}{|p{2cm}|p{1.5cm}|p{2cm}|p{2cm}|p{5cm}|p{2.5cm}|}
\hline
  \textbf{Name} & \textbf{Language} & \textbf{Pretrained from} & \textbf{Pretraining tasks} & \textbf{Corpus} & \textbf{Vocabulary} \\ [0.4cm] \hline
  
  BanglaBERT \cite{bhattacharjee2021banglabert} &	Bangla &	Scratch & 	RTD	 & Bangla Web text corpus &  	WordPiece (32k) \\ [0.4cm] \hline
  
  IndoBERT \cite{koto2020indolem} &	Indonesian &	Scratch &	MLM &	Indonesian Wikipedia, News and Web corpus	 & WordPiece (32k) \\ \hline
  
  IndonesianBERT \cite{wilie2020indonlu} &	Indonesian &	Scratch &	MLM	& Indo4B &	Sentencepiece (30k) \\ [0.4cm] \hline
  
  IndonesianBERT-Lite \cite{wilie2020indonlu} &	Indonesian &	Scratch & 	MLM  and SOP &	Indo4B &	Sentencepiece (30k) \\ [0.4cm] \hline
  
  PhoBERT \cite{nguyen2020phobert}	& Vietnamese &	Scratch &	MLM &	Vietnamese Wikipedia and  News corpus &	BPE (64K) \\ [0.4cm] \hline
  
RobBERT \cite{delobelle2020robbert} &	Dutch &	Scratch &	MLM	& OSCAR corpus &	bBPE (40K) \\ [0.4cm] \hline

RomanianBERT \cite{dumitrescu2020birth} & 	Romanian &	Scratch & 	MLM,NSP &	OPUS, OSCAR and Wikipedia corpus &	BPE (50k) \\ [0.4cm] \hline

FlauBERT \cite{le2020flaubert} & 	French	& Scratch &	MLM	& French text corpus &	BPE (50K) \\ [0.4cm] \hline 

AraBERT \cite{antoun2020arabert} & 	Arabic &	Scratch &	MLM and NSP & 	Arabic  Wikipedia and News corpus &	SentencePiece (64K) \\ [0.4cm] \hline

AraGPT2 \cite{antoun2021aragpt2} & 	Arabic &	Scratch &	CLM	& Arabic  Wikipedia, OSCAR and News corpus & 	bBPE (64K) \\ [0.4cm] \hline

AraELECTRA \cite{antoun2021araelectra} & 	Arabic &	Scratch & 	RTD & 	Arabic  Wikipedia, OSCAR and News corpus &	SentencePiece (64K) \\ [0.4cm] \hline

BERTje \cite{de2019bertje} &	Dutch &	Scratch & 	MLM, SOP &	Dutch   Books, Wikipedia and News corpus &	SentencePiece (30K) \\ [0.4cm] \hline

FinnishBERT \cite{virtanen2019multilingual} &	Finnish & 	Scratch &	MLM, NSP &	Finnish  News, Online discussion and Common Crawl Corpus &	SentencePiece (50K) \\ [0.4cm] \hline

ALBERTO \cite{polignano2019alberto} &	Italian	& Scratch &	MLM,NSP &	Italian tweets corpus &	SentencePiece  (128K) \\ [0.4cm] \hline

PortugueseBERT \cite{souza2020bertimbau}	& Portuguese &	mBERT &	MLM,NSP & 	BrWaC \cite{wagner2018brwac}	& SentencePiece (30K) \\ [0.4cm] \hline 

RuBERT \cite{kuratov2019adaptation}	& Russian &	mBERT	& MLM,NSP &	Russian Wikipedia and News corpus &	SentencePiece \\ [0.4cm] \hline

BETO \cite{canete2020spanish} &	Spanish &	Scratch &	MLM	& Wikipedia and Common Crawl corpus &	SentencePiece (32K) \\ [0.4cm] \hline

CamemBERT \cite{martin2020camembert} &	French &	Scratch &	MLM	& French OSCAR corpus &	SentencePiece(32K) \\ [0.4cm] \hline

ARBERT \cite{abdul2020arbert} &	Arabic &	Scratch & 	MLM, NSP &	Arabic Wikipedia, Books, Common Crawl, News corpus &	WordPiece (100K) \\ [0.4cm] \hline

MARBERT \cite{abdul2020arbert} &	Arabic &	Scratch	& MLM	& Arabic tweets corpus &	WordPiece (100K) \\ [0.4cm] \hline

WangchanBERTa \cite{lowphansirikul2021wangchanberta} & 	Thai &	Scratch &	MLM	& Thai Wikipedia , Social media posts, Books and reviews corpus & 	SentencePiece (25K) \\ [0.4cm] \hline

KoreALBERT \cite{lee2021korealbert} &	Korean &	Scratch &	MLM, SOP and WOP &	Korean Wikipedia, News, Internet crawl and Books corpus	& SentencePiece (32K) \\ [0.4cm] \hline

\end{tabular}}
\end{center}
\caption{\label{table-monolingual-1}  Summary of monolingual T-PTLMs.} 
\end{table*}

 A summary of the various monolingual model is presented in Tables \ref{table-monolingual-1} and \ref{table-monolingual-2}. Monolingual models are pretrained based on standard model architectures like GPT \cite{antoun2021aragpt2} BERT \cite{kuratov2019adaptation,souza2020bertimbau,wilie2020indonlu,koto2020indolem,le2020flaubert,abdul2020arbert,rybak2020klej,park2021klue,dumitrescu2020birth,antoun2020arabert,de2019bertje,virtanen2019multilingual,polignano2019alberto,malmsten2020playing,dadas2020pre,farahani2020parsbert}, RoBERTa \cite{park2021klue,canete2020spanish,nguyen2020phobert,delobelle2020robbert,martin2020camembert,lowphansirikul2021wangchanberta,straka2021robeczech}, ALBERT \cite{wilie2020indonlu,lee2021korealbert},  ELECTRA \cite{bhattacharjee2021banglabert,antoun2021araelectra}, and T5 \cite{carmo2020ptt5}. As the availability of corpus from one source is limited in the case of many languages, most of these models are pretrained using corpus gathered from multiple sources. For example, IndoBERT \cite{koto2020indolem} is pretrained on corpus having text from Wikipedia, News domain, and Internet. Except for models like PTT5 \cite{carmo2020ptt5}, RuBERT \cite{kuratov2019adaptation} and PortugueseBERT \cite{souza2020bertimbau}, all other monolingual models are pretrained from scratch. Models like PTT5, RuBERT, and PortugueseBERT are initialized from existing models and further pretrained with new language-specific vocabulary. In these models, only the transformer encoder layer parameters are copied from existing models while embedding layer parameters are randomly initialized. During CPT, embedding layer parameters are updated along with other layers. 
 
 \begin{table*}[t!]
\begin{center}
{\renewcommand{\arraystretch}{1.5}
\begin{tabular}{|p{2.4cm}|p{1.5cm}|p{1.6cm}|p{2cm}|p{5cm}|p{2.5cm}|}
\hline
  \textbf{Name} & \textbf{Language} & \textbf{Pretrained from} & \textbf{Pretraining tasks} & \textbf{Corpus} & \textbf{Vocabulary} \\ [0.4cm] \hline
  PTT5 \cite{carmo2020ptt5} &	Portuguese	& T5 &	Seq2SeqLM &	BrWac corpus &	SentencePiece (32K) \\ [0.4cm] \hline
  
SweedishBERT \cite{malmsten2020playing} &	Sweedish &	Scratch	& MLM ,NSP	& Sweedish text corpus (Wikipedia, News, Social media and  Legal) &	SentencePiece (50K) \\ [0.4cm] \hline

SweedishALBERT \cite{malmsten2020playing} &	Sweedish &	Scratch &	MLM, SOP &	Sweedish text corpus (Wikipedia, News, Social media and  Legal) &	SentencePiece (50K) \\ [0.4cm] \hline

SweedishELECTRA \cite{malmsten2020playing} &	Sweedish &	Scratch &	RTD	& Sweedish text corpus (Wikipedia, News, Social media and  Legal)	& SentencePiece (50K) \\ [0.4cm] \hline

HerBERT \cite{rybak2020klej}	& Polish &	Scratch &	MLM	& Polish text corpus (Wiki, OSCAR, Subtitles and Books)
&	BPE (50K) \\ [0.4cm] \hline

PolishBERT \cite{dadas2020pre} &	Polish	& Scratch	& MLM	& Polish text corpus (Common Crawl, Wikipedia, Books and OPUS)
	& SentencePiece (50K) \\ [0.4cm] \hline

RobeCzech \cite{straka2021robeczech}	& Czech	& Scratch &	MLM	& Wikipedia, Web and News corpus &	bBPE (52K) \\ [0.4cm] \hline

KLUE-BERT \cite{park2021klue} & 	Korean &	Scratch &	MLM,NSP &	Korean text corpus	& BPE(32K) \\ [0.4cm] \hline

KLUE-RoBERTa \cite{park2021klue} &	Korean &	Scratch	& MLM &	Diverse Korean text corpus	& BPE(32K) \\ [0.4cm] \hline

ParsBERT \cite{farahani2020parsbert} &	Persian	& Scratch &	MLM,NSP &	Persian text corpus	& WordPiece (100K) \\ [0.4cm] \hline

\end{tabular}}
\end{center}
\caption{\label{table-monolingual-2}  Summary of monolingual T-PTLMs.} 
\end{table*}

\subsubsection{Domain-Specific Models}
Following the success of T-PTLMs in general domain, T-PTLMs in specific domains like Finance \cite{yang2020finbert}, Legal \cite{leivaditi2020benchmark,chalkidis2020legal}, News \cite{gururangan2020don}, Programming \cite{lu2021codexglue,feng2020codebert,ahmad2021unified,guo2020graphcodebert,phan2021cotext}, Dialogue \cite{wu2020tod}, Networking \cite{louis2020netbert}, Academic \cite{liu2021oag,beltagy2019scibert,peng2021mathbert} and Biomedical \cite{lee2020biobert,alsentzer2019publicly,gu2020domain,peng2019transfer} have been developed (refer Table \ref{taxonomy-domain-table} for a brief summary).  Models like BERT, RoBERTa, BART, and T5 are pretrained on general domain text. For a model to perform well on domain-specific datasets the model should have enough domain knowledge \cite{lee2020biobert,yang2020finbert}. These general domain models can not acquire enough domain knowledge just through fine-tuning. As a result, the performance of these models on domain-specific datasets is limited \cite{lee2020biobert,yang2020finbert}. The initial trend to develop domain-specific models is using continual pretraining i.e., initialize the model with any of the existing general domain models and further pretrain on a domain-specific corpus. For example, BioBERT \cite{lee2020biobert} is the first domain-specific BERT model developed using continual pretraining. Following BioBERT in biomedical domain, models like AleaseBERT \cite{leivaditi2020benchmark}, RoBERTa-News \cite{gururangan2020don}, GraphCodeBERT \cite{guo2020graphcodebert}, CoText \cite{phan2021cotext}, CodeGPT-adapted \cite{lu2021codexglue}, NetBERT \cite{louis2020netbert}, MathBERT \cite{peng2021mathbert}, TOD-BERT \cite{wu2020tod}, ClinicalBERT \cite{alsentzer2019publicly} and BluBERT \cite{peng2019transfer} have been developed using continual pretraining.

\begin{table*}[t!]
\begin{center}
{\renewcommand{\arraystretch}{1.5}
\begin{tabular}{|p{2.5cm}|p{2cm}|p{1.5cm}|p{2.8cm}|p{3.8cm}|p{2.5cm}|}
\hline
  \textbf{Name} & \textbf{Domain} & \textbf{Pretrained from} & \textbf{Pretraining tasks} & \textbf{Corpus} & \textbf{Vocabulary}   \\ [0.4cm]
  \hline 
 FinBERT \cite{yang2020finbert} &	Finance &	Scratch &	MLM + NSP &	Financial Communication Corpus &	WordPiece (31K) \\ [0.4cm] \hline
 
ALeaseBERT \cite{leivaditi2020benchmark} &	Legal &	ALBERT &	MLM	& Lease Agreements &	Same as ALBERT \\ [0.4cm] \hline

LegalBERT \cite{chalkidis2020legal} &	Legal &	Scratch &	MLM + NSP &	English Legal Text & 	Sentencepiece (31K) \\ [0.4cm] \hline

RoBERTa-News \cite{gururangan2020don} &	News &	RoBERTa &	MLM	& Real News corpus &	Same as RoBERTa \\ [0.4cm] \hline

CodeBERT \cite{feng2020codebert} &	Programming	& Scratch &	MLM+RTD &	CodeSearchNet &	WordPiece \\ [0.4cm] \hline

PLBART \cite{ahmad2021unified} &	Programming	& Scratch &	DAE	& Github and Stackoverflow corpus &	Sentencepiece (50K) \\ [0.4cm] \hline

GraphCodeBERT \cite{guo2020graphcodebert} &	Programming &	CodeBERT &	MLM, EP and NA &	CodeSearchNet &	Same as CodeBERT \\ [0.4cm] \hline

CoText \cite{phan2021cotext} &	Programming &	T5	& Seq2SeqLM &	CodeSearchNet and Github code &	Same as T5 \\ [0.4cm] \hline

CodeGPT \cite{lu2021codexglue} &	Programming	& Scratch &	CLM	& CodeSearchNet &	BPE (50K) \\ [0.4cm] \hline

CodeGPT-adapted \cite{lu2021codexglue} &	Programming &	GPT-2 &	CLM	& CodeSearchNet &	Same as GPT-2 \\ [0.4cm] \hline

NetBERT \cite{louis2020netbert} &	Networking &	BERT &	MLM	 & Computer Networking Corpus &	Same as BERT \\ [0.4cm] \hline

SciBERT \cite{beltagy2019scibert} &	Academic &	Scratch &	MLM and NSP &	Semantic Scholar &	WordPiece (30K) \\ [0.4cm] \hline

OAG-BERT \cite{liu2021oag} &	Academic &	Scratch	& MLM & 	OAG text corpus & 	WordPiece (44K) \\ [0.4cm] \hline

MathBERT \cite{peng2021mathbert} &	Academic &	BERT &	MLM,CCP and MSP &	Arxiv papers &	Same as BERT \\ [0.4cm] \hline

TOD-BERT \cite{wu2020tod} &	Dialogue &	BERT &	MLM and RCL &	Dialogue Corpus &	Same as BERT \\ [0.4cm] \hline

BioBERT \cite{lee2020biobert} &	Biomedical &	BERT &	MLM+NSP &	PubMed and PMC	& Same as BERT \\ [0.4cm] \hline

ClinicalBERT \cite{alsentzer2019publicly} &	Biomedical &	BERT &	MLM+NSP &	MIMIC-III &	Same as BERT \\ [0.4cm] \hline

BlueBERT \cite{peng2019transfer} &	Biomedical &	BERT &	MLM+NSP &	PubMed and MIMIC-III &	Same as BERT \\ [0.4cm] \hline

PubMedBERT \cite{gu2020domain} &	Biomedical &	Scratch &	MLM+NSP &	PubMed and PMC &	WordPiece \\ [0.4cm] \hline

\end{tabular}}
\end{center}
\caption{\label{taxonomy-domain-table}  Summary of domain-specific T-PTLMs.} 
\end{table*}
 
The main advantage of developing domain-specific models using continual pretraining is that the model converges faster as it is not trained from scratch and hence it is comparatively less expensive. However, as these models use the same vocabulary learned over general domain text, many of the domain-specific words are missing in the vocabulary. For example, the vocabularies of FinBERT and general BERT have 41\% of common tokens \cite{yang2020finbert}.  As a lot of domain-specific words are missing in the vocabulary, many of the domain-specific words are not represented properly which hinders model learning. The advantage of having domain-specific vocabulary is that even if the word is missing in the vocabulary, the word will be split into meaningful tokens. For example, the word actyeltransferase is split into [“ace”, “ty”, “lt”, “ran”, “sf”, “eras”, “e’] by BERT model whereas the same word is split into meaning tokens [“acetyl”, “transferase”] by SciBERT which has domain-specific vocabulary \cite{gu2020domain}. Models like PubMedBERT \cite{gu2020domain}, FinBERT \cite{yang2020finbert}, LegalBERT \cite{chalkidis2020legal}, CodeBERT \cite{feng2020codebert}, PLBART \cite{ahmad2021unified}, CodeGPT \cite{lu2021codexglue}, SciBERT \cite{beltagy2019scibert} and OAG-BERT \cite{liu2021oag} are pretrained from scratch.

\subsection{Architecture}
\label{architecture-sec}

Transformers, a novel self-attention model deep learning proposed by Vaswani et al. \cite{vaswani2017attention} consists of stack of both encoder and decoder layers. A T-PTLM can be pretrained using a stack of encoders or decoders or both. 

\subsubsection{Encoder-based} In general, an encoder-based T-PTLM consists of an embedding layer followed by a stack of encoder layers. For example, the BERT-base model consists of 12 encoder layers while the BERT-large model consists of 24 encoder layers \cite{devlin2019bert}. The output from the last encoder layer is treated as the final contextual representation of the input sequence. In general, encoder-based models like BERT \cite{devlin2019bert}, XLNet \cite{yang2019xlnet}, RoBERTa \cite{liu2019roberta}, ELECTRA \cite{clark2019electra}, ALBERT \cite{lan2019albert} and XLM-E \cite{chi2021xlm} are used in NLU tasks.

\subsubsection{Decoder-based} A decoder-based T-PTLM consists of an embedding layer followed by a stack of decoder layers. Here transformer decoder layer consists of only masked multi-head attention and feed-forward network layers. The multi-head attention module which performs encoder-decoder cross attention is removed. In general, decoder-based models like GPT-1 \cite{radford2018improving}, GPT-2 \cite{radford2019language} and GPT-3 \cite{brown2020language} are used in NLG tasks. 

\subsubsection{Encoder-Decoder based} Encoder-decoder based T-PTLMs are more suitable for sequence-to-sequence modeling tasks like Machine Translation, Text Summarization, etc. MASS \cite{song2019mass} is the first encoder-decoder based T-PTLM model. It is pretrained using Seq2SeqLM, an extension of MLM to encoder-decoder architectures. Following MASS, a number of encoder-decoder models like T5 \cite{raffel2019exploring}, mT5 \cite{xue2021mt5}, mT6 \cite{chi2021mt6}, BART \cite{lewis2020bart}, mBART \cite{liu2020multilingual}, PLBART \cite{ahmad2021unified}, PEGAUSUS \cite{zhang2020pegasus} and PALM \cite{bi2020palm} are proposed in the recent times.  For example, Models like MASS and BART use bidirectional encoder over corrupted text and lef-to-right auto regressive decoder to reconstruct the original text.  

\subsection{SSL}
\label{type-ssl-sec}
SSL is one of the key ingredients in building T-PTLMs. A T-PTLM can be developed by pretraining using Generative, Contrastive or Adversarial, or Hybrid  SSL.

\subsubsection{Generative SSL} Generative SSL helps the model to learn by predicting tokens. The different scenarios in generative SSL are a) predicting the next token based on current tokens (CLM) b) prediction the masked tokens (MLM and its variants like TLM, Seq2SeqLM) c) reconstructing the original text from the corrupted text (DAE). Some of the popular models developed using Generative SSL are GPT-1 \cite{radford2018improving}, GPT-2 \cite{radford2019language}, GPT-3 \cite{brown2020language} (based on CLM), RoBERTa \cite{liu2019roberta}, XLM \cite{lample2019cross}, XLM-R \cite{conneau2020unsupervised} (based on MLM and its variants like TLM), BART \cite{lewis2020bart}, mBART \cite{liu2020multilingual} (based on DAE), and MASS \cite{song2019mass}, T5 \cite{raffel2019exploring}, mT5 \cite{xue2021mt5} (based on Seq2SeqLM).  

\subsubsection{Contrastive SSL} Contrastive SSL helps the model to learn by comparison. In NLP, there is no T-PTLM which is pretrained using Contrastive SSL only. Contrastive SSL is used in continual pretraining to further improve the model i.e., to learn sentence-level semantics. For example, CERT \cite{fang2020cert} uses contrastive SSL to improve the BERT model by injecting more sentence-level semantics. CERT outperforms BERT in many GLUE tasks. Similarly, Mirror-BERT \cite{liu2021fast} and SimCSE \cite{gao2021simcse} use contrastive SSL to allow the BERT model to generate quality sentence embeddings. Lin et al. \cite{lin2021common} showed that multi-lingual contrastive pretraining improves the performance of multilingual T-PTLMs in Mickey Probe. 

\subsubsection{Adversarial SSL} Adversarial SSL helps the model to learn by distinguishing corrupted tokens. Here the corrupted tokens can be replaced or shuffled. Adversarial SSL can be used in training the model from scratch or in continual pretraining.  Models like ELECTRA \cite{clark2019electra}  and XLM-E \cite{chi2021xlm} are pretrained using adversarial SSL. ELECTRA is pretrained using replaced token detection (RTD) while XLM-E is pretrained using multi-lingual replaced token detection(MRTD) and translation replaced token detection (TRTD). Panda et al. \cite{panda2021shuffled}  used adversarial SSL based on shuffled token detection (STD) to further improve RoBERTa model. 

\subsubsection{Hybrid SSL} Some of the T-PTLMs are pretrained using more than one type of SSL. For example, BERT model - generative (MLM) and contrastive SSL (NSP), ALBERT- generative (MLM) and contrastive SSL (SOP), infoXLM – generative (MLM, TLM) and contrastive SSL (XLCo). Here XLCo represents the cross lingual contrastive pretraining task. Models like CLINE \cite{wang2021cline} are obtained by further pretraining RoBERTa model using generative (MLM), contrastive, and adversarial SSL (RTD).

\subsection{Extensions}
\label{extensions-sec}

\subsubsection{Compact T-PTLMs}
PTLMs have achieved huge success in almost every NLP task.  Recently researchers observed that the performance of PTLMs can be increased just by increasing the size of the model and training with large volumes of corpus for more training steps \cite{kaplan2020scaling}. The large size and high latency make the deployment of PLTMs difficult in real-word applications where resources are limited and require fast inference. To reduce the size of T-PTLMs and make them faster, many model compression techniques like pruning, parameter sharing, knowledge distillation, and quantization are explored in recent times \cite{gupta2020compression}.

\textbf{Pruning}: In general, deep learning models like T-PTLMs are over parameterized i.e., some of the model components (weights \cite{gordon2020compressing}, attention heads \cite{michel2019sixteen,voita2019analyzing}, or layers \cite{sajjad2020poor,fan2019reducing} can be removed during pretraining or after pretraining without much impact on the model performance and also reducing the model storage space and inference time. Pruning is inspired from the biological observation, “thousands of trillions of synapses in a newborn baby reduces to 500 trillion synapses after ten years” \cite{gupta2020compression}. T-PTLMs are trained with multiple attention heads. Michel et al. \cite{michel2019sixteen} and Voita et al. \cite{voita2019analyzing} showed that most of the attention heads are redundant and can be removed during inference. Fan et al. \cite{fan2019reducing} showed that encoder layers can be dropped during pretraining which allows dropping layers during inference. In contrast, Sajjad et al. \cite{sajjad2020poor} applied layer dropping on the pre-trained models which eliminates training from scratch unlike Fan et al. \cite{fan2019reducing}.

\textbf{Knowledge Distillation}: Knowledge Distillation is a model compression method that allows training compact student models using the knowledge from large teacher models. During knowledge distillation, the student learns the generalization ability of the teacher model by reproducing its behavior, and hence the performance of the student model is on par with the teacher model. Knowledge Distillation is introduced by Bucila et al. \cite{bucilua2006model} and later generalized by Ba and Caruna \cite{ba2014deep} and Hinton et al. \cite{hinton2015distilling}.  The approach of Ba and Caruna   \cite{ba2014deep} trains the student model using L2 loss between teacher and student model logits while the approach of Hinton et al. \cite{hinton2015distilling} uses cross-entropy between softmax logits of teacher and student (soft loss) as well as cross-entropy loss between student prediction and actual label (hard loss). Some of the popular models trained using Knowledge distillation are DistilBERT \cite{sanh2019distilbert}, Tiny-BERT \cite{jiao2020tinybert}, BERT-PKD \cite{sun2019patient}, MobileBERT \cite{sun2020mobilebert}, and MiniLM \cite{wang2020minilm}. 

\textbf{Quantization}: Quantization compresses a model by using fewer bits to represent weights. In general, T-PTLMs parameters are represented using 32 or 16 bits. Quantized T-PTLMs use 8 bits \cite{zafrir2019q8bert} or even lesser \cite{shen2020q,zhang2020ternarybert,zadeh2020gobo} to present weights. Pruning compresses a model by removing less important weights, while quantization compresses a model by using fewer bits for weights representation. Some of the popular quantized BERT models are Q8BERT \cite{zafrir2019q8bert}, Q-BERT \cite{shen2020q}, and TernaryBERT \cite{zhang2020ternarybert}. To reduce performance drop in ultra-low (1 or 2) bit models, researchers proposed methods like mixed-bit quantization \cite{shen2020q,zadeh2020gobo}, combining knowledge distillation with quantization \cite{zhang2020ternarybert}, and product quantization (PQ) \cite{fan2020training}.  Mixed-bit quantization is not supported by some hardware while PQ requires extra clustering operations. Overall, quantization compresses the model by using fewer bits but as quantization is hardware specific, we need specialized hardware to use quantized models. 

\textbf{Parameter Sharing}: ALBERT \cite{lan2019albert} a lite version of the BERT model achieves parameter reduction by cross-layer parameter sharing and factorized embedding parameterization. Factorized embedding parameterization splits the large vocabulary matrix into two small matrices which allow growing the hidden vector size without significantly increasing vocabulary matrix parameters. Cross-layer parameter sharing prevents the growth of parameters with the increase in the depth of the model. With these two parameter reduction techniques, the ALBERT model has 18x fewer parameters and can be trained 1.7x faster compared to the BERT-large model. Cross-layer parameter sharing is also explored in Universal Transformers \cite{dehghani2018universal}.

\subsubsection{Character-based T-PTLMs}
Most of the T-PTLMs use sub-word embeddings based on tokenizers like BPE, bBPE, WordPiece, Unigram, and SentencePiece. The problem with the use of word embeddings is the requirement of a large vocabulary and OOV problem. Sub-word embeddings are based on the idea that only rare and misspelled words should be represented using sub-words while frequently used words should be represented as it is.  Sub-word embeddings overcome the two problems in word embeddings. However sub-word embeddings have two drawbacks a) cannot encode fine-grained character level information in the word representation and b) brittleness to noise i.e., even simple typos can change the representation of a word which hinders the model learning \cite{el2020characterbert,ma2020charbert}. 

To overcome these drawbacks, character-based T-PTLMs like CharacterBERT \cite{el2020characterbert}, CharBERT \cite{ma2020charbert}, and AlphaBERT \cite{chen2020modified} are proposed. CharacterBERT uses CharCNN+Highway layer to generate word representations from character embeddings and then apply transformer encoder layers. The use of CharCNN+Highway layer is inspired from ELMo \cite{peters2018deep}. Different from CharacterBERT which uses only character embeddings, CharBERT uses both character and sub-word embeddings.  In CharBERT, character-level word embeddings are generated from character embeddings using Bidirectional GRU similar to contextual string embeddings \cite{akbik2018contextual}. Then a dual-channel CNN is used in every transformer encoder layer to model the interaction between character and sub-word embedding channels. Due to the inclusion of an extra channel for character embeddings and interaction module in every layer, the size of the model increases by 5M parameters. CharBERT or CharRoBERTa are more robust to noise and perform better than BERT or RoBERTa models.  Unlike CharacterBERT and CharBERT, AlphaBERT in Biomedical domain operates directly at the character level. In AlphaBERT, transformer encoder layers are directly applied on character embeddings after adding to position embeddings.

\subsubsection{Green T-PTLMs} The standard approach to adapt general models to a specific domain or improve general models with knowledge from Knowledge bases is continual pretraining. The resulting models after continual pretraining achieve good results. But this process is expensive in terms of hardware and run time and also not environmentally friendly with CO2 emissions \cite{minot2021interpretable,strubell2019energy}. Recently, researchers focused on developing less expensive methods to adapt general models to a specific domain or to inject knowledge from knowledge bases. The models developed using less expensive methods like GreenBioBERT \cite{poerner2020inexpensive}, exBERT \cite{tai2020exbert}, and E-BERT \cite{poerner2020bert} are referred to as Green models as they are developed in a environmentally friendly way. 

GreenBioBERT \cite{poerner2020inexpensive} is developed by extending the vocabulary of the general BERT model using domain-specific word embeddings developed using Word2Vec which are further aligned with WordPiece embeddings. GreenBioBERT achieves comparable performance with BioBERT which is developed by further pretraining for 10 days using eight v100 NVIDIA GPUs. exBERT \cite{tai2020exbert} is developed by extending general BERT with domain-specific WordPiece embeddings and an extension module. During continual pretraining, as the extra WordPiece embeddings and extension module parameters are only updated while keeping other parameters freezed, this process is less expensive. E-BERT is developed by extending BERT model vocabulary with the Wikipedia2Vec entity vectors \cite{yamada2016joint} after aligning. E-BERT  \cite{poerner2020bert} which doesn’t require any further pretraining outperforms models like ERNINE \cite{zhang2019ernie} and KnowBERT \cite{peters2019knowledge} (both these models require further pretraining to inject information from knowledge bases) on the LAMA benchmark.  

\subsubsection{Sentence-based T-PTLMs}
The sentence embeddings obtained from T-PLTMs like BERT by applying any of the pooling strategies are not effective \cite{reimers2019sentence}. Recently many approaches based on supervised learning \cite{reimers2019sentence,cheng2021dual} or self-supervised learning \cite{zhang2020unsupervised,liu2021fast,wang2021tsdae,carlsson2020semantic,gao2021simcse,yan2021consert,carlsson2020semantic} are proposed. SBERT \cite{reimers2019sentence} is one of the first supervised approaches which extend T-PTLMs like BERT to generate quality sentence embeddings. SBERT fine-tunes BERT model using the Siamese network over NLI and STSb datasets. By fine-tuning over NLI and STSb which are sentence pair classification tasks, the model learns sentence-level semantics and hence generates quality sentence vectors. DvBERT \cite{cheng2021dual} which is based on multi-view learning \cite{xu2013survey} extends SBERT by adding word-level interaction features across two sentences. As supervised approaches require labeled datasets which limits the application of these models in labeled data scare domains. 

Moreover,  Zhang et  al. \cite{zhang2020unsupervised} showed that the performance of NLI and STSb fine-tuned models is limited when training data is limited or distribution of test differs significantly from training data. To overcome the requirement of labeled datasets, many approaches based on SSL are proposed. IS-BERT \cite{zhang2020unsupervised} uses mutual information maximization strategy to learn quality sentence embeddings and CNN instead of mean pooling on the top of BERT model. Mirror-BERT \cite{liu2021fast} trains BERT by using a contrastive learning-based objective to push positive sentence pairs to have similar representations. Here positive sentences are obtained by random masking in input space or applying dropout in feature space.  TSDAE \cite{wang2021tsdae} involves reconstructing the original sentence from the sentence embedding generated by the encoder using a corrupted sentence. SimCSE \cite{gao2021simcse} is a contrastive learning-based framework to further train PTLMs to generate better sentence embeddings using unlabelled or labeled data. 

\subsubsection{Tokenization-Free T-PLTMs} 
Most of the existing PTLMs use sub-word or character or both the embeddings. The main drawback with these approaches are 
\begin{itemize}
    \item Sub-word embeddings require a fixed vocabulary that is modest in size in models like BERT and RoBERTa but larger in multilingual models. The vocabulary requires a vocabulary matrix in which each token is mapped with a vector and softmax matrix in the output layer. These two matrix parameters occupy a significant amount of model parameters. For example, these two matrix parameters are about 66\% of mT5 model parameters \cite{xue2021mt5}. Moreover, having a fixed vocabulary makes the adaptation of models to other domains inefficient i.e., many domain-specific words are not represented properly which impacts model adaptation as well as model downstream performance \cite{el2020characterbert,gu2020domain}. 
    \item Both sub-word and character embeddings require an explicit tokenizer that splits the input sequence based on white space or punctuation. This becomes problematic in the case of languages that do not use white space or punctuations as word separators. For example, languages like Chinese and Thai do not use white space as separators and languages like Hawaiian and Twi use punctuations as consonants \cite{clark2021canine}. 
\end{itemize}

 Recently there is a rising interest in the research community to overcome the above drawbacks with tokenization-free T-PTLMs \cite{clark2021canine,xue2021byt5,tay2021charformer}. With tokenization-free models, there is no need for language-specific tokenizers, models are more robust to noise and have no large vocabulary which requires a significant amount of model parameters. CANINE \cite{clark2021canine} is the first tokenization-free T-PTLM which directly operates on character sequence. The model applies convolution layers on the character sequence to reduce the input sequence length and then applies transformer encoder layer stack. CANINE is pretrained with the same tasks as the BERT model. CANINE with 28\% fewer parameters compared to mBERT output performs it by 2.8 points on multilingual QA. 
 
 ByT5 \cite{xue2021byt5} is an improved version of T5 model to handle input at byte-level without using any fixed vocabulary. T5 uses the same number of encoder and decoder layers while the depth of the encoder is 3x compared to the decoder in ByT5. Charformer \cite{tay2021charformer} uses a novel tokenizer that uses gradients to automatically learn sub-words from characters which eliminate the requirement of a fixed vocabulary. Charformer performs on par with models like T5 while outperforming byte-level models like ByT5. Moreover,unlike CANINE, the gradient-based sub-word tokenizer output is interpretable. 

\subsubsection{Large Scale T-PTLMs}
Initially the sizes of T-PTLMs are in the range of 110M to 340M parameters \cite{devlin2019bert,liu2019roberta,clark2019electra}. Kaplan et al. \cite{kaplan2020scaling} showed that the performance of T-PTLMs is strongly related to the scale rather than the depth or width of the model. The authors showed that the performance of T-PTLMs is largely determined by the scale i.e., the number of parameters, the size of pretraining data, and the amount of pretraining compute. According to Kalplan et al. \cite{kaplan2020scaling} the performance of the model can be increased by increasing the size of the model or training the model on much large volumes of large data or training the model for more training steps. All these three must be scaled up at the same time to achieve optimal performance. This observation triggered the development of large-scale T-PTLMs like GPT-3 (175B) \cite{brown2020language}, PANGU(200B) \cite{zeng2021pangu}, GShard (600B) \cite{lepikhin2020gshard} which contains billions of parameters and Switch-Transformers (1.6T) \cite{fedus2021switch} which contains trillions of parameters.

\begin{figure*}[h]
\begin{center}
\includegraphics[width=17cm, height=8cm]{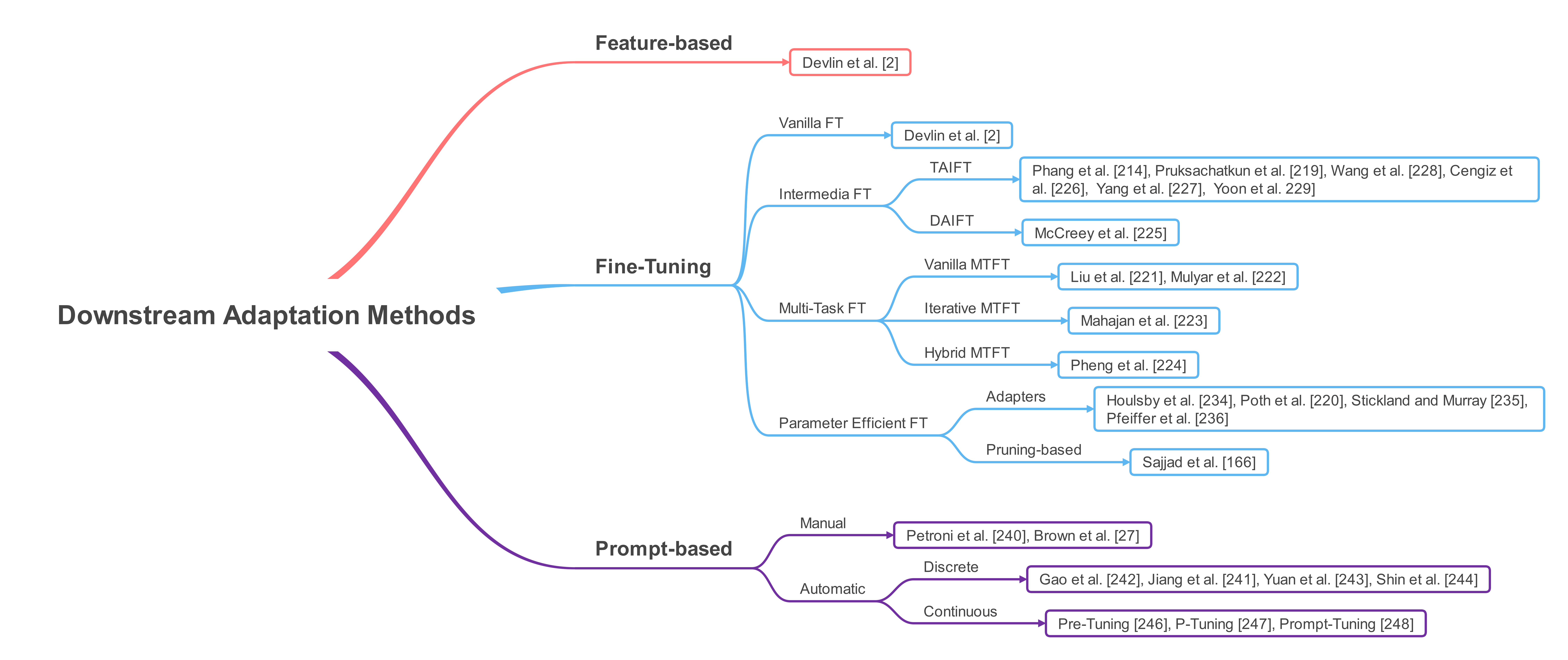}
\caption{\label{adaptation-methods} Downstream adaptation methods } 
\end{center}
\end{figure*}

\subsubsection{Knowledge Enriched T-PTLMs}
T-PTLMs are developed by pretraining over large volumes of text data. During pretraining, the model learns knowledge available in the pretraining text data by solving one ore more challenging pretraining tasks. Recent works \cite{li2021causalbert,lauscher2020specializing,zhou2020limit,ke2019sentilare,zhang2019ernie,hao2020enhancing,liu2021self,yuan2020coder,peters2019knowledge,levine2020sensebert} showed that these models can be further improved by integrating the knowledge available in external knowledge sources. T-PTLMs integrated with knowledge from external sources are referred to as Knowledge Enriched T-PTLMs.  Some of the popular external knowledge sources are WordNet, Wikidata in the general domain and UMLS \cite{bodenreider2004unified} in specific domains like Biomedical. Some of the examples of knowledge enriched T-PTLMs are CasualBERT \cite{li2021causalbert}, KnowBERT \cite{peters2019knowledge}, SenseBERT \cite{levine2020sensebert}, LIMIT-BERT \cite{zhou2020limit}, LiBERT \cite{lauscher2020specializing}, SentiLARE \cite{ke2019sentilare}, ERNINE \cite{zhang2019ernie}, E-BERT \cite{poerner2020bert} in the general domain and Clincal Kb-BERT \cite{hao2020enhancing}, Clinical Kb-ALBERT \cite{hao2020enhancing}, SapBERT \cite{liu2021self}, Sap-XLMR \cite{liu2021self}, UmlsBERT \cite{michalopoulos2020umlsbert}, CoderBERT \cite{yuan2020coder}, CoderBERT-All \cite{yuan2020coder} in specific domain like Biomedical. For example, CasualBERT injects casual knowledge into BERT model using two novel pretraining tasks on cause-effect pairs. LiBERT is pretrained from scratch using lingual relation classification (LRC) task along with MLM and NSP. LRC task helps to inject linguistic knowledge. LiBERT outperforms BERT model in most of the GLUE tasks. SentiLARE introduces label-aware MLM to inject POS tag and word polarity information into BERT model. All the biomedical domain-specific knowledge-enriched models are obtained by integrating knowledge from the biomedical ontology UMLS. 

\subsubsection{Long-Sequence T-PTLMs}
The self-attention attention module in transformers updates the representation of each input token by attending to all tokens in the input sequence. The quadratic time complexity of the self-attention module limits the application of T-PTLMs to long input sequences. To overcome this drawback, self-attention variants like sparse self-attention and linearized self-attention are proposed to reduce its complexity and hence extend T-PTLMs to long input sequences also \cite{lin2021survey}. Some of the popular T-PTLMs based on a) sparse self-attention are Longformer \cite{beltagy2020longformer}, ETC \cite{ainslie2020etc}, BigBird \cite{zaheer2020big} and Reformer \cite{kitaev2019reformer} and b) linearized self-attention are Performer \cite{choromanski2020rethinking}. Sparse self-attention reduces the complexity by including sparsity bias which reduces the number of query-key pairs that each query attends to. In linearized self-attention, reduced complexity is achieved by disentangling the attention with kernel feature maps and then computing the attention in reverse order.  

\subsubsection{Efficient T-PTLMs}
T-PTLMs require pretraining on large volumes of text data for longer durations which makes pretraining highly expensive. Recently, with better model architectures it is possible to achieve similar or better performances using less pretraining data \cite{he2020deberta} and less pretraining costs \cite{jiang2020convbert}. DeBERTa \cite{he2020deberta} improves the BERT model using disentangled attention mechanism and enhanced masked decoder. Disentangled attention mechanism represents a word using separate vectors to encode its content and position information and then compute the attention weights based on contents and relative positions.  An enhanced masked decoder is used to predict masked tokens instead of softmax layer during pretraining. These two novel changes improve pretraining efficiency and DeBERTa model which is pretraining on 78GB of data outperforms RoBERTa which is pretrained on 160GB of data. 

ConvBERT \cite{jiang2020convbert} improves the BERT model with mixed attention block consisting of self-attention and span based dynamic convolution modules. The self-attention modules model global dependencies while span-based dynamic convolution modules model local dependencies. The authors of ConvBERT observed that some attention heads are needed to model only local dependencies and hence they replaced these attention heads with span-based dynamic convolution modules.  ConvBERT with better model architecture outperforms ELECTRA base using less than ¼ of its pretraining cost. 

\section{Downstream Adaptation Methods}
\label{adaptation-methods-sec}
Once a language model is pretrained, it can be used in downstream tasks. A pretrained language model can be used in downstream tasks in three ways namely a) feature-based b) fine-tuning and c) prompt-based tuning (refer Figure \ref{adaptation-methods}). The feature-based approach involves generating contextual word embeddings from language models and then using them as input features in task-specific downstream models. Fine-tuning involves adapting model weights to downstream tasks by minimizing task-specific loss.

\subsection{Feature-based} In traditional deep learning models like CNN or RNN, word embeddings generated using embedding models like Word2Vec \cite{mikolov2013efficient} or Glove \cite{pennington2014glove} are used as word features. In feature-based approach, BERT \cite{devlin2019bert} based models are used to generate contextual word vectors, and then they are used as input features similar to Word2Vec or Glove embeddings in task-specific downstream models. BERT-based contextual word embeddings are much better as a) they are contextual unlike Word2Vec and Glove embeddings b) overcome the issue of OOV words and c) encode more information in word vectors because of the deep layered model architecture. Here, word vectors can be taken from the last layer (or from multiple layers using any of the pooling strategies) \cite{devlin2019bert}. The advantage with the feature-based approach is that contextualized word vectors can be used any in any of the handcrafted state-of-the-art task specific-architectures. However, feature-based approach involves training the downstream model from scratch (except embeddings) which requires a large number of labeled instances. 

\subsection{Fine-tuning} Pretraining allows the pretrained language model to gain universal language knowledge. However, the performance of the model in downstream tasks requires task-specific knowledge i.e., for the model to perform well in downstream tasks, its weights should be close to the ideal setting for the target task \cite{phang2018sentence}. Fine-tuning imparts task-specific knowledge to the model by adapting its weights based on task-specific loss \cite{devlin2019bert}. Moreover, fine-tuning enhances the model performance because it clusters the points of different labels away from each other such that there is a large separation between the cluster regions 
\cite{zhou2021closer}. Fine-tuning updates all the transformer layers including the embedding layer but the higher layers are subjected to more changes compared to the lower layers \cite{zhou2021closer,merchant2020happens,mosbach2020interplay,hao2020investigating}. 

Models like BERT, RoBERTa, and ELECTRA do not follow unified input-output format across tasks i.e., different tasks have different input and output formats. So, it is required to add task-specific layers in these models during fine-tuning. However, in models like T5 \cite{raffel2019exploring} which follow the same input-output format across tasks, there is no need to add any extra layers specific to each task. T5 follows a text-to-text format in any task i.e., input for the model is some text and the model has to produce some text as output. 

Fine-tuning can be a) Vanilla fine-tuning \cite{devlin2019bert} b) Intermediate fine-tuning \cite{phang2018sentence,pruksachatkun2020intermediate,poth2021pre} c) Parameter efficient fine-tuning and d) Multi-task fine-tuning \cite{liu2019multi,mulyar2020mt,mahajan2020identification,peng2020empirical}. Unlike Vanilla fine-tuning which is prone to overfit the model on small datasets, intermediate fine-tuning or multi-task fine-tuning avoid overfitting the model on small datasets. As fine-tuning involves adjustments to the entire model weights, methods like adapters or pruning-based fine-tuning help to fine-tune the model in a parameter-efficient way.

 \subsubsection{Vanilla Fine-Tuning} In Vanilla fine-tuning, the model is adapted to downstream tasks based on task-specific loss \cite{devlin2019bert}. The main drawback in vanilla fine-tuning is that PTLM having large parameters is prone to overfit on small task-specific datasets. Moreover, with small datasets, the model weights are not adapted well to the end task which limits its performance. Intermediate fine-tuning or multi-task fine-tuning overcome the issues in vanilla fine-tuning. 

 \subsubsection{Intermediate Fine-Tuning (IFT)} IFT involves fine-tuning the model on an intermediate dataset with a large number of labeled instances. IFT helps the model to gain additional domain or task-specific knowledge which avoids overfitting and enhances its performance on small target datasets \cite{phang2018sentence,pruksachatkun2020intermediate,poth2021pre}.  Poth et al. \cite{poth2021pre} established that intermediate pre-training can yield performance gains in adapter-based setups, similar to what has been previously found for full model finetuning.  IFT can be domain adaptive \cite{mccreery2019domain} or task adaptive \cite{phang2018sentence,cengiz2019ku_ai,yang2020measurement,wang2020learning,yoon2019pre,jeong2020transferability}.
   
   \textit{Domain adaptive intermediate fine-tuning (DAIFT)}: DAIFT involves fine-tuning the model on the same domain dataset with a large number of labeled instances i.e., source and target datasets are of the same domain but different tasks. DAIFT on the same domain source dataset imparts more domain knowledge to the model which enhances the model performance on the same domain target task \cite{mccreery2019domain}. McCreery et al. \cite{mccreery2019domain} fine-tuned models like BERT and XLNet on the medical question-answer pairs dataset to enhance the performance on the Medical question similarity dataset. Here source (medical question-answer pair) dataset and target (medical question similarity) datasets are from the same domain i.e., Medical but from different tasks.  The models (BERT and XLNet) are pretrained on a general domain text corpus. DAFT on medical domain dataset injects medical knowledge into BERT and XLNet which are pretrained on a general domain text corpus. McCreery et. al. \cite{mccreery2019domain} showed that the improvement is more when the number of training instances in the target task is less.
   
   \textit{Task adaptive intermediate fine-tuning (TAIFT)}: TAIFT involves fine-tuning the model on the same or related task dataset with a large number of labeled instances i.e., source and target datasets are from the same or related task. Here the source and target datasets need not be from the same domain. TAIFT on the same or related task source dataset imparts more task-specific knowledge to the model which enhances the model performance on the target dataset. For example, Cengiz et al. \cite{cengiz2019ku_ai} showed that TAIFT on general domain NLI datasets improves the in-domain model performance on the medical NLI dataset. Similarly, Yang et al. \cite{yang2020measurement} and Wang et al. \cite{wang2020learning} achieved better performance on clinical STS dataset with TAIFT on general STS dataset. Yoon et al. \cite{yoon2019pre} showed that TAIFT on the general domain SQUAD dataset improves the performance on the biomedical question answering dataset. Jeong et al. \cite{jeong2020transferability} improved BioBERT model performance in biomedical QA with TAIFT on the general NLI dataset. Phang et al. \cite{phang2018sentence} achieved an improvement of 1.4 in GLUE score for BERT with TAIFT on the general NLI dataset. Further, the authors showed that the improvement is more when target labeled instances are less in number. TAIFT on NLI datasets imparts sentence-level reasoning skills to the model which improves the model performance in other tasks. 
   
However, IFT does not guarantee better performance all the time \cite{phang2018sentence,pruksachatkun2020intermediate,wang2019can} i.e., IFT sometimes negatively impacts the transferability to downstream tasks. Pruksachatkun et al. \cite{pruksachatkun2020intermediate} performed a large-scale study on the pretrained RoBERTa model with 110 intermediate–target task combinations to investigate when and why IFT is beneficial. The authors showed that intermediate tasks requiring high-level inference and reasoning abilities tend to work best i.e., NLI and QA tasks that involve common sense reasoning are generally useful as intermediate tasks.

\subsubsection{Multi-task Fine-Tuning (MTFT)} Multi-task learning (MTL) allows the model to learn knowledge that is useful across tasks. The primary focus of MTL can be improving the performance of target tasks with the help of auxiliary tasks or improving the performance of all the tasks \cite{worsham2020multi}. The advantages of MTL are a) allows the model to gain more knowledge by learning from multiple datasets simultaneously which reduces the requirement of a large number of labeled instances in a specific target task. b) provides a regularization effect by avoiding overfitting to a specific target task \cite{liu2019multi}. Multi-task fine-tuning can be a) Vanilla MTFT b) Iterative MTFT and c) Hybrid.
  
  \textit{Vanilla MTFT} : Vanilla MTFT involves fine-tuning the model on multiple datasets simultaneously \cite{liu2019multi,khan2020mt}. For example, Liu et al. \cite{liu2019multi} improved the performance of BERT model in GLUE tasks using vanilla MTFT. Here, the embedding and transformer layers are shared across the tasks while each task has a task-specific layer. However, it is not guaranteed that Vanilla MTFT always improves the performance of the model across tasks \cite{mulyar2020mt}. For example, Mulyar et al. \cite{mulyar2020mt} developed MT-ClinicalBERT by fine-tuning ClinicalBERT using multiple datasets related to NER, STS, and RTE tasks. MT-ClinicalBERT achieved on par but less performance compared to task-specific ClinicalBERT models. The possible reason for this is that some of the tasks may negatively transfer knowledge which reduces the performance of the model. The drawback in vanilla MTFT can be avoided using Iterative MTFT which allows selecting the best set of tasks or MTFT followed by vanilla fine-tuning. 
  
  \textit{Iterative MTFT}: Iterative MTFT allows to select the best set of tasks for fine-tuning the model \cite{mahajan2020identification}. It is necessary to select the best set of related datasets as vanilla MTFT on all the tasks sometimes may degrade the performance of the model \cite{mulyar2020mt}. Iterative MTFT is similar to traditional feature selection in machine learning. Iterative MTFT helps to select the best set of datasets to fine-tune the model whereas feature selection helps to select the best set of features. Mahanjan et al. \cite{mahajan2020identification} applied iterative MTFT to choose the best set of related datasets and achieved SOTA results on the clinical STS dataset. 
  
  \textit{Hybrid MTFT}: Iterative MTFT allows to choose the best set of related datasets, but it is expensive as it involves multiple iterations. Moreover, each iteration involves training the model on multiple datasets and then fine-tuning it on the target dataset.  Instead of iteratively applying MTFT, we can fine-tune the model on multiple related datasets and then fine-tune it on the target dataset with a small learning rate \cite{peng2020empirical}.  We refer to this as hybrid MTFT as it involves vanilla MTFT followed by vanilla fine-tuning. 
  
\begin{figure*}[h!]
\begin{center}
\includegraphics[width=18cm, height=7cm]{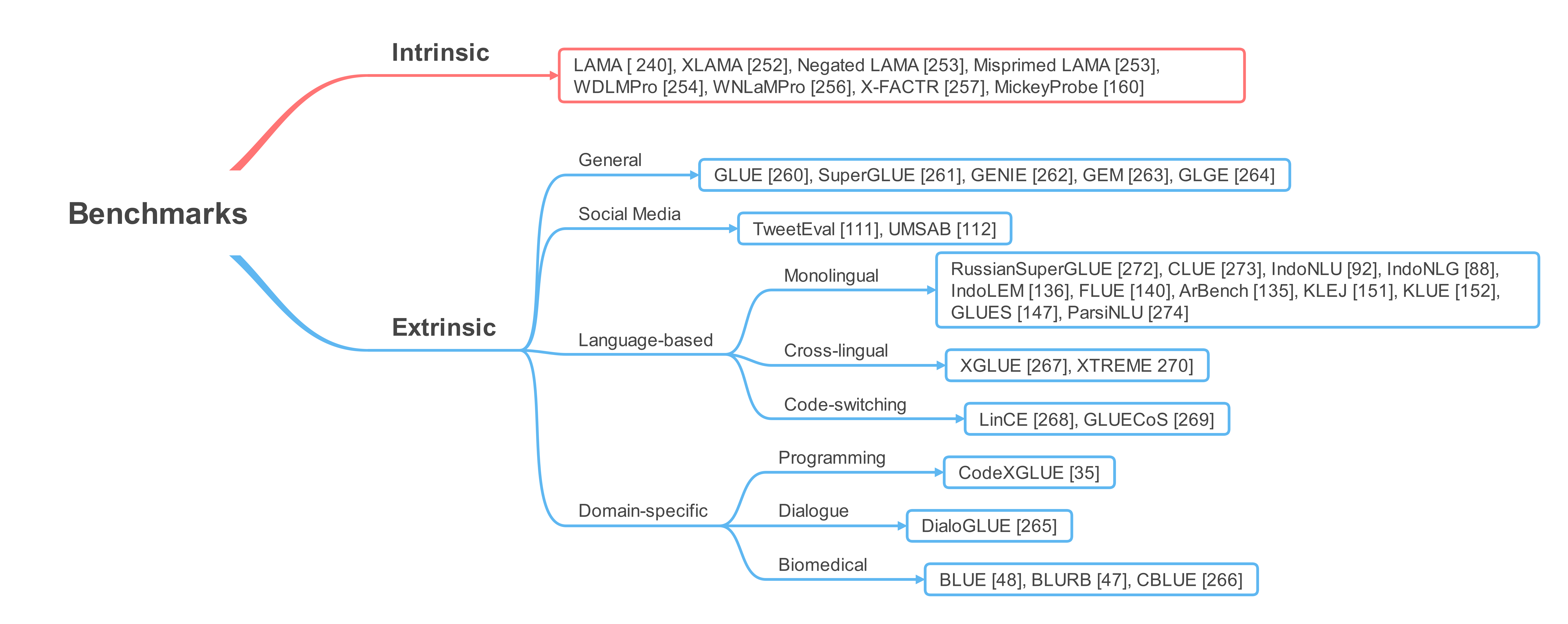}
\caption{\label{benchmarks} Benchmarks to evaluate the progress in T-PTLMs} 
\end{center}
\end{figure*}
  
\subsubsection{Parameter Efficient Fine-Tuning} Fine-tuning allows the model weights to adapt to downstream tasks by minimizing the task-specific loss i.e., fine-tuning starts with copying the entire model weights and making small changes. As fine-tuning involves updating the entire model weights, it is required to train a separate model for each task which is not parameter efficient. Adapters \cite{houlsby2019parameter} and pruning-based fine-tuning \cite{sajjad2020poor} helps to fine-tune the model in a parameter-efficient way. 
  
  \textit{Adapters \cite{houlsby2019parameter}}– The adapter is a special trainable layer module proposed by Houlsby et al. \cite{houlsby2019parameter} to fine-tune pretrained language models in a parameter-efficient way. The adapter module consists of two feed-forward layers with a non-linear layer in between and a skip connection. The adapter module projects the input vector into a small vector and then projects back into the original dimension using the two feed-forward layers and non-linear layer. Let $x$ be the original vector dimension and y be the small vector dimension, then the total number parameters in the adapter module are $2xy + x+ y$. By setting $x<<y$, we can further reduce the number of parameters in the adapter module. The small vector dimension ($y$) provides a trade-off between performance and parameter efficiency. Adapters are added to each of the sublayers in transformer layer before layer normalization. During fine-tuning, only parameters of adapters, layer normalization in each transformer layer, and task-specific layers are only updated while the rest of the parameters in pretrained model are kept frozen. 
  
Houlsby et al. \cite{houlsby2019parameter} showed that adapter-based fine-tuning is highly parameter efficient and they can achieve the performance of a fully fine-tuned model using adapter-based fine-tuning which involves only 3\% of task-specific parameters. Moreover, Poth et al. \cite{poth2021pre} showed that intermediate fine-tuning using adapters improve model performance in the target task. Stickland and Murray \cite{stickland2019bert} proposed an approach to train adapters in a multi-task setting. However, this approach suffers from issues like a) requirement of simultaneous access to multiple datasets and b) difficulty in balancing various tasks as the model may overfit on low resource tasks and underfit on high resource tasks. Pfeiffer et al. \cite{pfeiffer2020low} proposed AdapterFusion a novel two-stage method based on adapters that overcome the issues in sequential learning and multi-task learning to leverage knowledge from multiple tasks. They showed that AdapterFusion outperforms full fine-tuning as well the adapter-based model trained in single and multi-task setups.

  \textit{Pruning-based fine-tuning} - Recent studies \cite{kovaleva2019revealing,michel2019sixteen,voita2019analyzing,dalvi2020analyzing,sajjad2020poor} show that deep pre-trained language models have redundancy. Pruning methods are based on the idea that not all the parameters are important in the pre-trained models and some of them can be removed without much impact on the model performance. For example, research studies \cite{kovaleva2019revealing,michel2019sixteen,voita2019analyzing} show that some of the attention heads can be pruned. Sajjad et al. \cite{sajjad2020poor} proposed different strategies to drop encoder layers in pretrained language models. The authors showed that the size of the pre-trained models can be reduced by dropping encoder layers after pre-training and the resulting pruned BERT model can be fine-tuned to get competitive performance. The experimental results show that it is possible to prune BERT, RoBERTa, and XLNet models by up to 40\% while maintaining up to 98\% of their original performance.

\subsection{Prompt-based Tuning}
In general, most of the P-TLMs are pretrained using language modeling objectives and then adapted
to downstream tasks using fine-tuning which involves task-specific objectives. The discrepancy in objectives during pretraining and fine-tuning impacts the downstream performance of the model.  The downstream performance of models can be improved especially in few-short and zero-shot settings by prompt-based tuning which formulates the tuning process as a slot filling which is close to the language modeling objective. Here the prompt can be close or pre-fix shape and it can be generated manually or automatically \cite{liu2021pre}. Close style prompts are suitable for models pretrained masked language modeling objective while pre-fix style prompts are suitable for models pretrained using casual modeling objective. 

Prompt-based tuning initially is based on manually created prompts. For example, LAMA \cite{petroni2019language} probe is conducting manually created close-style prompts while GPT-3 \cite{brown2020language} model is tuned using manually created pre-fix style prompts. As manually creating prompts is a time-taking process and these prompts can be sub-optimal also \cite{jiang2020can}. To over these drawbacks prompts are created automatically. Automatically generated prompts can be discrete or continuous. A discrete prompt is simply a string i.e., a sequence of words included in the input text to guide the T-PTLM to better model the downstream task. Some of the popular methods to generate discrete prompts are Prompt mining \cite{jiang2020can}, Prompt generation \cite{gao2020making}, Prompt paraphrasing \cite{jiang2020can,yuan2021bartscore}, and Gradient-based search \cite{shin2020eliciting,wallace2019universal}. 
              
Prompt mining \cite{jiang2020can} involves collecting a large number of sentences having the subject $x$ and object $y$ and then generate the new prompts using the middle words or the dependency paths. Prompt generation \cite{gao2020making} involves generating prompts using T-PTLMs like T5. Prompt paraphrasing generates new prompts from seed prompts using methods like back translation \cite{jiang2020can} or equivalent phrases from thesaurus \cite{yuan2021bartscore}. Gradient-based search generates trigger tokens which can be combined with input sequence to create a prompt \cite{shin2020eliciting,wallace2019universal}. 

Continuous prompts perform prompting in the embedding space of T-PTLM  i.e., add a sequence of task-specific vectors to the input sequence. Here the prompt vectors need not be embeddings of natural language words. Unlike discrete prompts where the template parameters are determined by T-PTLM parameters, in continuous prompts, the templates have their parameters independent of T-PTLM parameters. Some of the popular continuous prompt generating approaches are Pre-tuning \cite{li2021prefix}, P-Tuning \cite{liu2021gpt}, and Prompt-tuning \cite{lester2021power}. Further Lester et al. \cite{lester2021power} showed that prompt ensembling outperforms traditional model ensembling. Here prompt ensembling means generating multiple prompts for the same task.

\section{Evaluation}
\label{evaluation}
A T-PTLM gains knowledge encoded in pretraining corpus during pretraining. Here the knowledge refers to syntactic, semantic, factual, and common-sense knowledge. The effectiveness of a T-PTLM can be evaluated in two ways namely intrinsic and extrinsic (refer Figure \ref{benchmarks}). Intrinsic evaluation probes the knowledge encode in T-PLTM while extrinsic evaluation evaluates how effective the T-PTLMs are in real-world downstream tasks. Intrinsic evaluation sheds light on the knowledge gained by T-PTLM during pretraining which helps us to design better pretraining tasks so that the model learns more knowledge during the pretraining stage itself.

\subsection{Intrinsic Evaluation}
\begin{table*}[t!]
\begin{center}
{\renewcommand{\arraystretch}{1.5}
\begin{tabular}{|p{2cm}|p{3cm}|p{2cm}|p{1.5cm}|p{3cm}|p{3.5cm}|}
\hline
  \textbf{Name} & \textbf{Probe} & \textbf{Language} & \textbf{Method} & \textbf{Includes} &  \textbf{Data Source} \\
  \hline
  LAMA \cite{petroni2019language} & 	Factual and Common-sense knowledge &	English &	UnTQ &  	Single token entities &	TREx \cite{elsahar2019t}, GoogleRE, ConceptNet \cite{speer2012representing} and SQUAD \cite{rajpurkar2016squad} \\ \hline
  
XLAMA \cite{kassner2021multilingual}	& Factual knowledge	& Multilingual (53 languages) &	TQ (Typed Query) &	Single and Multi token entities &	TREx \cite{elsahar2019t} and GoogleRE \\ \hline

Negated LAMA \cite{kassner2020negated} &	Impact of negation in probing factual and common-sense knowledge &	English	& UnTQ &	Single token entities &	TRex \cite{elsahar2019t}, GoogleRE, ConceptNet \cite{speer2012representing}and SQUAD \cite{rajpurkar2016squad}  \\ \hline

Misprimed LAMA \cite{kassner2020negated} &	Impact of mis primes in probing factual and common-sense knowledge &	English &	UnTQ &	Single token entities &	TRex \cite{elsahar2019t}, GoogleRE, ConceptNet \cite{speer2012representing}and SQUAD \cite{rajpurkar2016squad} \\ \hline

WDLMPro \cite{senel2021does} &	Word Understanding &	English	& Ranking &	Single and Multi-token entities &	WordNet \cite{fellbaum2010wordnet} \\ \hline

WNLaMPro \cite{schick2020rare} &	Word Understanding &	English &	UnTQ &	Single and Multi-token entities &	WordNet \cite{fellbaum2010wordnet} \\ \hline

X-FACTR \cite{jiang2020x} &	Factual knowledge &	Multilingual (26 languages) &	UnTQ &	Single and Multi-token entities &	TREx \cite{elsahar2019t} \\ \hline

MickeyProbe \cite{lin2021common} &	Common-sense knowledge &	Multilingual (11 languages) &	Sentence ranking	& - &	OMCS Corpus \cite{singh2002open} \\
\hline
  
\end{tabular}}
\end{center}
\caption{\label{table-intrinsic}  Summary of various intrinsic benchmarks.} 
\end{table*}

Intrinsic evaluation involves probing the model knowledge using probes like LAMA \cite{petroni2019language}, XLAMA \cite{kassner2021multilingual}, X-FACTR \cite{jiang2020x}, MickeyProbe \cite{lin2021common} , Negated LAMA \cite{kassner2020negated}, Misprimed LAMA \cite{kassner2020negated}, WDLMPro \cite{senel2021does} or WNLaMPro \cite{schick2020rare} (refer Table \ref{table-intrinsic}). LAMA is one of the first probes introduced to evaluate factual and common-sense knowledge in PLTMs under zero-shot settings. LAMA consists of a corpus of facts where the fact can be a relation triplet or a question-answer pair gathered from SQUAD. Here facts are converted to fill-in-the-blank style questions and the model is evaluated based on the prediction of blank tokens i.e., \textit {argmax x}$\in$\textit{W P(x⁄temp)} represents the model vocabulary and $temp$ represents the fill-in-the-blank template.  LAMA is based on the hypothesis that a model with a good amount of factual knowledge correctly predicts the blank tokens i.e., i.e., the ground truth tokens are predicted with the highest probability compared to other tokens in the model vocabulary. Negated LAMA and Misprimed LAMA probe shows that the language models are not able to consider the negated or misprimed words in the templates. For example, the model predicts the same token whether the templated is negated or not.  Poerner et al. (2020) \cite{poerner2020bert}  introduced LAMA-UHN which is a collection of triples from the LAMA probing benchmark which are difficult to guess. 

The main drawbacks in the LAMA probe are a) restriction to single token entities only b) limits the prediction of tokens over the model vocabulary which hinders the evaluation of models with different vocabulary c) it probes only English language models and d) many of the triples in LAMA are easy to guess \cite{poerner2020bert}. XLAMA extends the LAMA probe to multiple languages (53 languages) and includes multi-token entities also. Moreover, in LAMA the model has to predict over the entire model vocabulary while in XLAMA the model has to predict over a fixed set of candidates specific to each relation type i.e., \textit {argmax x}$\in$\textit{C P(x⁄temp)}    where $C$ represents a set of candidate entities specific to a relation type. The authors of XLAMA refer to this type of querying as Typed Query (TQ) and the querying in LAMA as UnTyped Query (UnTQ). Similar to XLAMA, X-FACTR is also a multi-lingual probe for 23 languages. Moreover, the authors of X-FACTR developed several decoding algorithms to predict multi-token entities. MickeyProbe \cite{lin2021common} is a zero-shot common-sense probe that uses sentence-level ranking based on Pseudo-Likelikhood \cite{salazar2020masked}. Here the model has to rank a set of declarative sentences having similar words and syntactic features. The performance of multilingual models in retrieving knowledge varies with language i.e., it is high in high resource languages compared to low resource languages \cite{kassner2021multilingual,jiang2020x,lin2021common} . Moreover, the multilingual model exhibits language bias i.e., the language of query affects the model prediction \cite{kassner2021multilingual}. The model performance in retrieving knowledge can be improved by pretraining on cod-switched data \cite{jiang2020x} or further pretraining using a multilingual contrastive loss function \cite{lin2021common}. 

Probes like LAMA, XLAMA, and X-FACTR focus on evaluating the relations between entities. Unlike these probes, WDLMPro and WNLaMPro focus on understanding how the pretrained models understand the words. WNLamPro uses fill-in-the-black style templates while WDLMPro evaluates the model by matching a word with its definition. WDLMPro probe is based on the assumption that a model correctly matches a word with its definition only when the model understands the word. WDLMPro consists of synset groups where each group consists of a word, its taxonomic sister words from WordNet along their definitions. 

\subsection{Extrinsic Evaluation}
\begin{table*}[t!]
\begin{center}
{\renewcommand{\arraystretch}{1.5}
\begin{tabular}{|p{1.8cm}|p{1.5cm}|p{2.5cm}|p{1.5cm}|p{1.3cm}|p{1.3cm}|p{5cm}|}
\hline
  \textbf{Benchmark} & \textbf{Type} & \textbf{Category} & \textbf{Language} & \textbf{Public Leaderboard} & \textbf{Diagnostic Dataset} &  \textbf{Details} \\ \hline
  
  GLUE \cite{wang2018glue} &  NLU	& General &	English &  \cmark &  \cmark & 	Five NLU tasks (TC, SA, STS, PI, and NLI) with nine datasets and one diagnostic dataset. \\ \hline 
  
  SuperGLUE \cite{wang2019superglue} &	NLU	& General &	English & \cmark &  \cmark &	Five NLU tasks (QA, WSD, NLI, and Coref) with eight datasets and two diagnostic datasets. \\ \hline
  
  GENIE \cite{khashabi2021genie} &  NLG &	General &	English &  \cmark &  \xmark &	Four NLG tasks (MT, TS, MRC, and CSR). \\ \hline
  
  GEM \cite{gehrmann2021gem} & 	NLG &	General &	English & \cmark &  \xmark &	Four NLG tasks (Data2text, TS, TSim, and Dialog) with thirteen data sets. \\ \hline
  
  GLGE \cite{liu2020glge} &	NLG	 & General	& English &  \cmark &  \xmark &	Eight language generation tasks, including Abstractive TS, Answer-aware Question Generation, Conversational QA, and Personalizing Dialogue. \\ \hline
  
  TweetEval \cite{barbieri2020tweeteval} & NLU &	Social-media &	English & \cmark &  \xmark &	Seven tweets related tasks, and all are framed as multi-class tweet classification. \\ \hline
  
  UMSAB \cite{barbieri2021xlm} &	XLU &	Social-media &	Cross-lingual &	 \cmark &  \xmark & Cross-lingual sentiment analysis for eight different languages and all are framed as tweet classification with three labels (positive, negative, and neutral). \\ \hline 
  
  CodeXGLUE \cite{lu2021codexglue} & 	PLU and PLG	& Domain-specific &	Programming & \cmark &  \xmark &	Ten programming language understanding and generation tasks with fourteen datasets. \\ \hline
  
  DialoGLUE \cite{mehri2020dialoglue} & 	NLU &	Domain-specific (Dialogue) &	English & \cmark &  \xmark &	Four NLU tasks (Intent Prediction, Slot-filling, Semantic parsing, and Dialogue state tracking) with seven task-oriented dialogue datasets. \\ \hline

  BLUE \cite{peng2019transfer} &	NLU &	Domain-specific (biomedical) & 	English		&	\cmark &  \xmark & Five tasks (STS, NER, RE, NLI and DC) with ten datasets that cover both biomedical and clinical texts with different dataset sizes and difficulties. \\ \hline

  BLURB \cite{gu2020domain} &	NLU	& Domain-specific (biomedical) &	English	&	\cmark &  \xmark & Thirteen biomedical NLP datasets in 6 tasks (NER, PICO, RE, STS, DC, and QA). \\ \hline

  CBLUE \cite{zhang2020conceptualized} &	NLU	& Domain-specific (biomedical) &	Chinese	& \cmark &  \xmark & Includes tasks like NER, PI, QA, IR, IC, and ToC.
\\ \hline

\end{tabular}}
\end{center}
\caption{\label{table-extrinsic-1}  Summary of general, social-media and domain-specific Extrinsic Benchmarks. TC - Text Classification, STS- Semantic Text Similarity, TS - Text Summarization, TSim – Text Simplification, ToC – Topic Classification, IC – Intent Classification, IR – Information Retrieval, QA- Question Answering, DC- Document Classification, RE – Relation Extraction, POS - Parts-of-speech tagging, DP - Dependency Parsing, MRC - Machine Reading Comprehension, SA – Sentiment Analysis, PI- Paraphrase Identification, NLI – Natural Language Inference, WPR – Web Page Ranking, QAM – Question Matching, QADSM – Query Ad Matching, MT- Machine Translation, LID – Language Identification, CSR – Common Sense Reasoning, WSD – Word Sense Disambiguation, MCQA – Multiple Choice Question Answering.} 
\end{table*}

Extrinsic evaluation helps to assess the performance of a model in downstream tasks. To get the maximum out of a model, the model should perform well across a wide range of tasks rather than just performing well on one or two tasks. A benchmark provides a standard way of evaluating the model’s generalization ability across tasks. A benchmark usually consists of a set of datasets, a leader board, and a single metric \cite{wang2018glue}. The datasets are chosen in a way that they are challenging and represent diverse tasks. A leaderboard is an online repository that helps to compare and rank models. For a model to achieve a good score in a benchmark, it should share knowledge i.e., parameters across tasks with one or two layers specific to each task \cite{wang2018glue}. A benchmark uses a single metric to evaluate the overall performance of the model across tasks. Without a benchmark, it is difficult to evaluate models in a standard way and track the progress in the development of pretrained language models. A summary of various extrinsic bechmarks are presented in Tables \ref{table-extrinsic-1} and \ref{table-extrinsic-2}.

\begin{table*}[t!]
\begin{center}
{\renewcommand{\arraystretch}{1.5}
\begin{tabular}{|p{1.8cm}|p{1.5cm}|p{2.5cm}|p{1.5cm}|p{1.3cm}|p{1.3cm}|p{5cm}|}
\hline
  \textbf{Benchmark} & \textbf{Type} & \textbf{Category} & \textbf{Language} & \textbf{Public Leaderboard} & \textbf{Diagnostic Dataset} &  \textbf{Details} \\ \hline
  
  XGLUE \cite{liang2020xglue} & 	XLU and XLG & 	Language-based &	Cross-lingual & \cmark &  \xmark &	Eleven tasks in which nine tasks are XNLU (NER, POS, QA, NLI, PI, WPR, QAM, QADSM and TC) and two tasks are XNLG (question and news title generation). This benchmark covers 19 languages. \\ \hline
  
  LinCE \cite{aguilar2020lince} & 	NLU and NLG &	Language-based &	Code-Switching & \cmark &  \xmark &	Five tasks (MT, LID, NER, POS, and SA) with eighteen datasets covering nine different code-switched language pairs. \\ \hline
  
  GLUECoS \cite{khanuja2020gluecos} & NLU &	Language-based &	Code-Switching & \cmark &  \xmark &	Eleven datasets covering six tasks (LID, POS, NER, SA, QA and NLI) and two language pairs (English-Hindi and English-Spanish). \\ \hline
  
  XTREME \cite{hu2020xtreme} & XLU &	Language-based	& Cross-lingual & \cmark &  \xmark &	Nine tasks spanning forty typologically diverse languages from 12 language families. \\ \hline
  
  XTREME-R \cite{ruder2021xtreme} &	XLU	& Language-based	& Cross-lingual	& \cmark &  \cmark & Includes ten challenging NLU tasks for 50 languages.\\ \hline
  
  RussianSuper GLUE \cite{shavrina2020russiansuperglue} &	NLU &	Language-based &	Russian &  \cmark &  \xmark &	Nine Russian NLU tasks. \\ \hline

  IndicGLUE \cite{kakwani2020inlpsuite} &	NLU &	Language-based	& Indian languages & \cmark &  \xmark &	Ten tasks covering multiple Indian languages. \\ \hline

   CLUE \cite{xu2020clue} &	NLU &	Language-based &	Chinese & \cmark &  \cmark &	Nine language understanding tasks in Chinese and diagnostic dataset for linguistic analysis. \\ \hline

   IndoNLU \cite{wilie2020indonlu} &	NLU	& Language-based &	Indonesian & \cmark &  \xmark &	Twelve tasks clustered into four categories: (a) single-sentence classification, (b) single-sentence sequence tagging, (c) sentence-pair classification, and (d) sentence-pair sequence labelling. \\ \hline

   IndoNLG \cite{cahyawijaya2021indonlg} &	NLG &	Language-based	& Indonesian &	\xmark &  \xmark & Six commonly used NLG tasks: TS, QA, Chitchat, and three different pairs of machine translation (MT) tasks.  \\ \hline

   IndoLEM \cite{koto2020indolem} &	NLU &	Language-based &	Indonesian &  	\cmark &  \xmark &	Seven tasks for the Indonesian language spanning Morpho-syntax and Sequence labelling, Semantics, and Discourse with eight datasets.  \\ \hline

   FLUE \cite{le2020flaubert} &	NLU and XLU &	Language-based & French &	\xmark &  \xmark &	Six tasks (TC, PI, NLI, WSD, DP, and POS). Three out of six tasks (TC, PI, and NLI) are from cross-lingual datasets. \\ \hline

  ArBench \cite{abdul2020arbert} &	NLU &	Language-based &	Arabic &	\cmark &  \xmark &	Six different Arabic language understanding tasks (SA, Social meaning tasks, ToC, Dialect identification, NER, and QA) with 42 datasets. \\ \hline

  KLEJ \cite{rybak2020klej} &	NLU	& Language-based &	Polish	&	\cmark &  \xmark & Seven tasks (NER, Semantic relatedness, QA, TE, SA, and Cyberbully detection) with 9 datasets.  \\ \hline

  KLUE \cite{park2021klue} &	NLU &	Language-based &	Korean	&	\cmark &  \xmark & Eight Korean natural language understanding tasks, including ToC, STS, NLI, NER, RE, DP, MRC, and Dialogue state tracking.  \\ \hline

  GLUES \cite{canete2020spanish} & 	NLU	& Language-based &	Spanish	&	\xmark &  \xmark & Includes tasks like NLI, PI, NER, POS, DC, DP and QA.  \\ \hline
  
  ParsiNLU \cite{khashabi2020parsinlu} &	NLU	& Language-based & Persian &	\cmark &  \xmark &	Includes six NLU tasks like TE, PI, SA, MT, MRC, MCQA. \\ \hline
 
\end{tabular}}
\end{center}
\caption{\label{table-extrinsic-2}  Summary of language-based  extrinsic benchmarks. TC - Text Classification, STS- Semantic Text Similarity, TS - Text Summarization, TSim – Text Simplification, ToC – Topic Classification, IC – Intent Classification, IR – Information Retrieval, QA- Question Answering, DC- Document Classification, RE – Relation Extraction, POS - Parts-of-speech tagging, DP - Dependency Parsing, MRC - Machine Reading Comprehension, SA – Sentiment Analysis, PI- Paraphrase Identification, NLI – Natural Language Inference, WPR – Web Page Ranking, QAM – Question Matching, QADSM – Query Ad Matching, MT- Machine Translation, LID – Language Identification, CSR – Common Sense Reasoning, WSD – Word Sense Disambiguation, MCQA – Multiple Choice Question Answering.} 
\end{table*}
GLUE \cite{wang2018glue} and SuperGLUE \cite{wang2019superglue} benchmarks are the commonly used benchmarks to evaluate the natural language understanding ability of pretrained language models. GLUE benchmark consists of nine tasks which include both single sentence and sentence pair tasks. With rapid progress in model development, the models achieved good performance in the GLUE benchmark resulting in little space for further improvement \cite{wang2019superglue}. To have a more challenging benchmark, the SuperGLUE benchmark is introduced with more challenging tasks like QA, word sense disambiguation (WSD), and coreference resolution while retaining the two difficult tasks from GLUE benchmark. 

Inspired by the success of GLUE and SuperGLUE benchmarks in the general English domain, benchmarks like GENIE \cite{khashabi2021genie}, GEM \cite{gehrmann2021gem}, GLGE \cite{liu2020glge} have been introduced to evaluate NLG models in the general English domain.  To evaluate cross-lingual models, XGLUE \cite{liang2020xglue} and XTREME \cite{hu2020xtreme} benchmarks have been introduced. XTREME benchmark includes only XNLU tasks while the XGLUE benchmark includes both XNLU and XNLG tasks. Moreover, the XGLUE benchmark includes diverse datasets related to search, ads, and news scenarios which makes it more challenging and practical. Recently, as there is less room for improvement in the XTREME benchmark with existing achieving improvements by almost 13 points, Ruder et al. \cite{ruder2021xtreme} extended XTREME to XTREME-R which consists of ten challenging NLU tasks. Moreover, XTREME covers only forty languages while XTREME-R covers 50 languages.  

To evaluate social media-based T-PTLMs, we have benchmarks like TweetEval \cite{barbieri2020tweeteval} and UMSAB \cite{barbieri2021xlm}. TweetEval includes datasets from English only while UMSAB includes datasets from eight languages including English. In both the benchmarks, all the tasks are framed as tweet classification. Apart from XGLUE and XTREME which evaluate cross-lingual models, we have separate benchmarks in each language like Russian (RussianSuperGLUE \cite{shavrina2020russiansuperglue}), Indian (IndicGLUE \cite{kakwani2020inlpsuite}), Chinese (CLUE \cite{xu2020clue}), Indonesian (IndoNLU \cite{wilie2020indonlu}, IndoNLG \cite{cahyawijaya2021indonlg}, IndoLEM \cite{koto2020indolem}), French (FLUE \cite{le2020flaubert}), Arabic (ArLUE \cite{abdul2020arbert}), Polish (KLEJ \cite{rybak2020klej}), Korean (KLUE \cite{park2021klue}) Spanish (GLUES \cite{canete2020spanish}), and Persian (ParsiNLU \cite{khashabi2020parsinlu})  to evaluate monolingual language models. Besides, we have benchmarks like GLUECoS \cite{khanuja2020gluecos} and LinCE \cite{aguilar2020lince} for CodeSwitching, BLUE \cite{peng2019transfer}, BLURB \cite{gu2020domain} and ChineseBLUE \cite{zhang2020conceptualized} in the Biomedical domain, CodeXGLUE \cite{lu2021codexglue} in the Code intelligence domain and DialogGLUE \cite{mehri2020dialoglue} to evaluate Dialog models. Further, we have benchmarks like FewCLUE \cite{xu2021fewclue}, FLEX \cite{bragg2021flex}, and FewGLUE \cite{schick2021s} to evaluate T-PTLMs under few shot settings. 

\section{Useful Libraries}
\label{libraries-sec}
We present a summary of popular libraries to work with transformer-based PTLMs. Libraries like Transformers \cite{wolf2020transformers} and Fairseq \cite{ott2019fairseq} are useful for model training and evaluation. Some of the libraries like SimpleTransformers, HappyTransformer, AdaptNLP which are built on the top of Transformers library make the model training and evaluation easier with just a few lines of code. Libraries like FastSeq \cite{yan2021fastseq}, DeepSpeed \cite{rasley2020deepspeed}, FastT5, OnnxT5 and LightSeq \cite{wang2021lightseq} are useful to increase the inference speed of models. Ecco, BertViz \cite{vig2019bertviz}, and exBERT \cite{hoover2020exbert} are visual analysis tools to explore the layers of transformer models while Transformers-interpret and Captum help to explain the model decisions. 

\begin{table*}[p]
\begin{center}
\begin{tabular}{|p{2.5cm}|p{2cm}|p{5cm}|p{2cm}|p{5cm}|}
\hline
  \textbf{Library}  &   \textbf{Purpose}    &   \textbf{Description}     &    \textbf{Framework}   &  \textbf{Link} \\
  \hline 
Transformers \cite{wolf2020transformers} &  Training and Inference &	State-of-the-art library for transformer based PTLMs. &	Pytorch, Tensorflow and Jax & 	\url{https://github.com/huggingface/transformers} \\ \hline 

SimpleTransformers & Training and Inference &	Built on the top of transformers and lets you to quickly train and evaluate models. &	PyTorch &	\url{https://github.com/ThilinaRajapakse/simpletransformers} \\ \hline 

HappyTransformer & Training and Inference &	Built on the top of transformers and makes the use of state-of-the-art models easy. &	PyTorch &	\url{https://github.com/EricFillion/happy-transformer} \\ \hline 

FairSeq \cite{ott2019fairseq} & Training and Inference	 & Library to train custom models for translation, summarization, language modeling and other text generation tasks. &	PyTorch &	\url{https://github.com/pytorch/fairseq} \\ \hline 

AdaptNLP & Training and Inference &	Built on the top of Flair and Transformers library and  makes the use of state-of-the-art models easy. &	PyTorch &	\url{https://github.com/Novetta/adaptnlp} \\ \hline

SimpleT5 & Training and Inference &	Built on top of PyTorch-lightning and Transformers that lets you quickly train your T5 models. &	PyTorch-Lightning &	\url{https://github.com/Shivanandroy/simpleT5} \\ \hline 

SpacyTransformers & All NLP tasks &	spaCy pipelines for pretrained BERT, RoBERTa XLNet, GPT-2 etc. & PyTorch &	\url{https://github.com/explosion/spacy-transformers} \\ \hline 

TextBox & Text Generation &	Library for building text generation systems based on models like GPT-2, BART, T5 etc.	& PyTorch	& \url{https://github.com/RUCAIBox/TextBox} \\ \hline 

Trankit & Multilingual NLP &	Light-Weight Transformer-based Python Toolkit for Multilingual Natural Language Processing and is built on the top of transformers library.	& PyTorch &	 \url{https://github.com/nlp-uoregon/trankit} \\ \hline 

Haystack & Information Retrieval & Library to build powerful and production-ready pipelines for different search use cases. &	PyTorch &	\url{https://github.com/deepset-ai/haystack} \\ \hline 

EasyNMT & Machine Translation &	Easy to use, state-of-the-art Neural Machine Translation library for 100+ languages.	& PyTorch	& \url{https://github.com/UKPLab/EasyNMT} \\ \hline 

Aitextgen & Text Generation &	Library for training and generation using OpenAI's GPT-2 and EleutherAI's GPT Neo/GPT-3 architecture. &	PyTorch-Lightning &	\url{https://github.com/minimaxir/aitextgen} \\ \hline 

Dl-Translate & Machine Translation &	Deep Learning-based translation library built on Huggingface transformers. &	PyTorch &	\url{https://github.com/xhlulu/dl-translate} \\ \hline 

FastSeq \cite{yan2021fastseq} & Fast Inference &	Efficient implementation of the popular sequence models for text generation, summarization, and translation tasks. &	PyTorch	& \url{https://github.com/microsoft/fastseq} \\ \hline 

LightSeq \cite{wang2021lightseq} & Fast Inference	& High performance training and inference library for sequence processing and generation. &	PyTorch, Tensorflow	& \url{https://github.com/bytedance/lightseq} \\ \hline 

TurboTransformers \cite{fang2021turbotransformers} & Fast Inference &	A library open source by WeChat AI to get fast inference using transformer models.	& PyTorch &	\url{https://github.com/Tencent/TurboTransformers} \\ \hline 

EET & Fast Inference &	PyTorch library to make transformer models inference faster. &	PyTorch &	\url{https://github.com/NetEase-FuXi/EET} \\ \hline 

DeepSpeed \cite{rasley2020deepspeed} & Distributed Model Training &	Deep learning optimization library that makes distributed training easy, efficient, and effective. &	PyTorch &	\url{https://github.com/microsoft/DeepSpeed} \\ \hline 

FastT5 & Fast Inference &	Reduce T5 model size by 3X and increase the inference speed up to 5X. &	PyTorch &	\url{https://github.com/topics/fastt5} \\ \hline 

OnnxT5 & Fast Inference	& Fast Inference of T5 model. &	PyTorch	& \url{https://github.com/abelriboulot/onnxt5} \\ \hline 

exBERT \cite{hoover2020exbert} & Visualization &	Library to explore the learned attention weights and contextual representations. &	PyTorch &	\url{https://github.com/bhoov/exbert} \\ \hline 

BertViz \cite{vig2019bertviz} & Visualizaation &	Library to visualize attention in the Transformer model. &	PyTorch &	\url{https://github.com/jessevig/bertviz} \\ \hline 

Transfomers-interpret & Model Interpretation &	Library to explain the decision of transformer models. &	PyTorch	 & \url{https://github.com/cdpierse/transformers-interpret} \\ \hline 

Ecco & Visualization &	Library to visualize and explore NLP language models. &	PyTorch &	\url{https://github.com/jalammar/ecco} \\ \hline 

Captum & Model Interpretation &	PyTorch interpretation library. &	PyTorch	& \url{https://github.com/pytorch/captum} \\ \hline 

TextBrewer \cite{yang2020textbrewer} & Model Compression &	Supports Knowledge distillation methods. &	PyTorch &	\url{https://github.com/airaria/TextBrewer} \\ \hline 

KD-Lib \cite{shah2020kd} & Model Compression &	Library to develop compact model using model compression techniques like quantization, pruning and knowledge distillation. &	PyTorch &	\url{https://github.com/SforAiDl/KD_Lib} \\ \hline 

Parallelformers	& Model Parallelization & An Efficient Model Parallelization Toolkit for Deployment (inference). &	PyTorch 	 & \url{https://github.com/tunib-ai/parallelformers} \\ 

 \hline
\end{tabular}
\end{center}
\caption{\label{libraries}  Useful Libraries to work with T-PTLMs} 
\end{table*}

\section{Discussions and Future Directions}
\label{future-directions}
\subsection{Better Pretraining Methods} It is highly expensive to pretrain a model especially large-scale models with billions or trillions of parameters using SSL only. Novel pretrained methods like Knowledge Inherited Pretraining (KIPT) involve both SSL and Knowledge Distillation \cite{qin2021knowledge}. SSL allows the model to learn the knowledge available in pretraining corpus while KD allows the model to learn the knowledge already encoded in existing pretrained models. Due to the additional knowledge gained by the model during pretraining through KD, a) the model converges faster and hence reduces the pretraining time b) the model performs better in downstream tasks compared to the models pretrained using SSL only \cite{qin2021knowledge}. The research community must focus more on developing better pretraining methods like KIPT which allow the model to gain more knowledge as well as reduce the pretraining time.

\subsection{Sample Efficient Pretraining Tasks} A pretraining task is sample efficient if it makes maximum out of each train instance i.e., it should be defined over all the tokens in the training instance. Sample efficient pretraining tasks make pretraining more compute efficient \cite{clark2019electra}.  MLM, the most commonly used pretraining task is less sample efficient as it involves only a subset of tokens i.e., masked tokens which amount to 15\% of total tokens \cite{devlin2019bert,clark2019electra}. Pretraining tasks like RTD \cite{clark2019electra}, RTS \cite{di2021efficient}, and STD \cite{panda2021shuffled} can be considered as early attempts to develop sample-efficient pretraining tasks. All these three pretraining tasks are defined over all the tokens in each training instance i.e., they involve identifying whether each token is replaced \cite{clark2019electra}, randomly substituted \cite{di2021efficient}, or shuffled \cite{panda2021shuffled} or not. We can expect more sample efficient pretraining tasks which make pretraining more compute efficient. 

\subsection{Efficient Models}
Pretraining T-PLMs is highly expensive due to the large model size and also there is a requirement of large volumes of unlabelled text data. However long pretraining times are not environmentally friendly due to CO2 emission and the availability of large volumes of unlabelled text data is not possible in all the domains like Biomedical. Recently, models like DeBERTa \cite{he2020deberta} with novel improvements to BERT model achieve better performance than RoBERTa model even though it is pretrained using just 78GB of data which is just half of the data used to pretrain RoBERTa model. Similarly, ConvBERT \cite{jiang2020convbert} with a novel mixed attention module outperforms ELECTRA model using just ¼ of its pretraining cost. There is a great need for efficient models like DeBERTa and ConvBERT to reduce the amount of pretraining data as well as the pretraining costs.

\subsection{Better Position Encoding Mechanisms}
The self-attention mechanism is permutation invariant without position bias. The position bias can be provided using absolute or relative position embeddings. Moreover, absolute position embeddings can be predetermined or learned. However, there are drawbacks to both these approaches \cite{likhomanenko2021cape}. Absolute position embeddings suffer from generalization issues but are easy to implement. Unlike absolute positions, relative position embeddings are robust to sequence length changes but difficult to implement and yield less performance.   There is a great need for more novel position encoding mechanisms like CAPE \cite{likhomanenko2021cape} which combines the advantages in both absolute and relative position embeddings. 

\subsection{Improving existing T-PTLMs} T-PTLMs like BERT and RoBERTa have achieved good results in many of the NLP tasks. Recent research works showed that these models can be further improved by injecting sentence-level semantics through continual pretraining based on adversarial \cite{panda2021shuffled} or contrastive pretraining tasks \cite{lin2021common,fang2020cert} . For example, Panda et al. \cite{panda2021shuffled} showed that continual pretraining using shuffled token detection objective improves RoBERTa model performance in GLUE tasks by allowing the model to learn more coherent sentence representations. Similarly, continual pretraining using contrastive pretraining objectives improves the performance of T-PTLMs in GLUE tasks \cite{fang2020cert} and multilingual T-PTLMs in Mickey Probe \cite{lin2021common}. Further research is required to extend this to other monolingual and domain-specific T-PTLMs.   

\subsection{Beyond Vanilla Fine-tuning} Fine-tuning is the most commonly used method to adapt pretrained models to downstream tasks. However, the main drawback with vanilla fine-tuning is that it makes changes to all the layers in the pretrained model, and hence it requires maintaining a separate copy for each task which makes deployment expensive.   Methods like Adapters \cite{houlsby2019parameter} and Pruning-based tuning \cite{sajjad2020poor} are proposed to adapt pretrained models to downstream tasks in a parameter-efficient way. For example, adapters are small task-specific layers added to each transformer layer and during downstream task adaptation, only adapter layer parameters are updated while keeping the transformer layer parameter fixed. Moreover, Poth et al. \cite{poth2021pre} showed that adapters are useful for intermediate fine-tuning also.  Recently prompt-based tuning methods (discrete – \cite{shin2020eliciting,gao2020making,brown2020language}  and continuous – \cite{li2021prefix,lester2021power}) have attracted the research community with much better parameter efficiency. For example, prompt-based tuning methods like Prefix-tuning \cite{li2021prefix} require only 0.1\% of task-specific parameters while adapter-based fine-tuning involves 3\% of task-specific parameters \cite{houlsby2019parameter}.  

\subsection{Benchmarks} In the last four layers many benchmarks have been introduced to evaluate the progress in pretrained models in general \cite{wang2018glue,wang2019superglue,khashabi2021genie,gehrmann2021gem,liu2020glge} as well as in specific domains \cite{lu2021codexglue,mehri2020dialoglue,barbieri2020tweeteval,peng2019transfer,gu2020domain,zhang2020conceptualized}. Apart from English, benchmarks are introduced to evaluate the progress in other monolingual \cite{shavrina2020russiansuperglue,xu2020clue,wilie2020indonlu,cahyawijaya2021indonlg,koto2020indolem,le2020flaubert,abdul2020arbert,rybak2020klej,park2021klue,canete2020spanish,khashabi2020parsinlu} as well as multilingual models \cite{liang2020xglue,hu2020xtreme,barbieri2021xlm}. However, the existing benchmarks are not sufficient to cover all the scenarios. For example, there are no benchmarks to evaluate a) the progress in compact pretrained models b) the robustness of pretrained models c) PTLMs specific to social media as well as specific to other domains like Academic. Recently,  leaderboards like Explainboard \cite{liu2021explainaboard} which not only evaluate the progress using a single metric like existing benchmarks but also dig deeper by analyzing the strengths and weaknesses of models are introduced. This kind of leaderboard should be extended to other domains also. Moreover, benchmarks like FewGLUE \cite{schick2021s}, FLEX \cite{bragg2021flex}, and FewCLUE \cite{xu2021fewclue} which evaluate few-shot learning techniques should be extended to other languages and domains also. 

\subsection{Compact Models} Transformer-based PTLMs achieved state-of-the-art results in almost every NLP task. However, these models are large which requires more amount of storage space. As these models have many layers through which the input has to pass through to get the model prediction, latency is high \cite{gupta2020compression}. As real-world applications are resource-constrained and require less latency, model compression methods like pruning, quantization, knowledge distillation, parameter sharing, and factorization are explored to develop compact models in English for general domain applications \cite{gupta2020compression,menghani2021efficient}. There is a great need to explore these model compression methods to develop compact models for other languages as well as for other domains also.

\subsection{Robustness to Noise} Transformer-based PTLMs are brittle to noise which includes both adversarial and natural noise \cite{sun2020adv,pruthi2019combating} . The main reason behind this is the use of sub-word embeddings. In the case of sub-word embeddings, even a small typo error can change the overall representation of the word by breaking the word into many sub-word tokens which hinders model learning and impact the model predictions \cite{el2020characterbert,ma2020charbert}. To increase the robustness of PTLMs to noise, models like CharacterBERT \cite{el2020characterbert} use character embeddings only while models like CharBERT \cite{ma2020charbert} use character embeddings along with sub-word embeddings. Both these approaches improved the robustness to noise. Recently, tokenization-free models like CANINE \cite{clark2021canine}, ByT5 \cite{xue2021byt5}, and Charformer \cite{tay2021charformer} are proposed which further improve robustness to noise. There is a need for more robust models to increase the use of PTLMs in real-world applications especially in sensitive domains like Medicine. 

\subsection{Novel Adaptation Methods} The commonly used strategy to adapt general models to specific domains like biomedical or multilingual models to specific languages is continual pretraining \cite{lee2020biobert,alsentzer2019publicly,peng2019transfer}. Although this approach achieves good results by adapting the model to a specific domain or language, the lack of domain or language-specific vocabulary hurts the model downstream performance. Recently researchers proposed methods like vocabulary expansion \cite{poerner2020inexpensive}, vocabulary expansion and then continual pretraining \cite{tai2020exbert,wang2020extending}. These methods overcome the issue of OOV words but increase the size of vocabulary due to the addition of new terms in the vocabulary. Recently, Yao et al. \cite{yao2021adapt} proposed the Adapt and Distill approach to adapt general models to a specific domain using vocabulary expansion and knowledge distillation. Different from existing adaptation methods, this approach not only adapts general models to specific domain but also reduces the size of the model. Further research on this topic will result in more novel adaption methods.

\subsection{Privacy Issues} Transformer-based PTLMs achieved impressive results in many of the NLP tasks. However, there are some unexpected as well as unwanted risks associated with these models. For example, data leakage from these models is of primary concern especially when the model is pretrained over private data. As the model is pretrained over a large amount of text data, it is possible to recover sensitive data including personally identifiable information \cite{nakamura2020kart,carlini2020extracting,misra2019black,hisamoto2020membership}. This prevents the public release of models pretrained on private data. Recently, Carlini et al. \cite{carlini2020extracting} showed that GPT-2 model generates the entire postal address of a person which is included in training data when prompted with the person’s name. Recently frameworks like KART \cite{nakamura2020kart} are introduced in the Biomedical domain which performs various attacks to assess data leakage. There is a great need to develop more sophisticated attacks to assess data leakage and also methods to prevent leakage of sensitive data from pretrained models.  

\subsection{Mitigating Bias} Deep learning-based models are increasingly used in many real-world applications including specific domains like Biomedical \cite{wang2021cloud} and Legal \cite{araci2019finbert}. However, these models are prone to learn and amplify the bias already present in training data. As a result, the decisions from these models are biased i.e., may favor a particular race, gender, or aged people. This behavior is completely undesirable. Some of the recent works focused on identifying and mitigating bias. For example, Minot et al. \cite{minot2021interpretable} proposed a data augmentation-based approach to reduce gender bias while Liang et al. \cite{liang2021towards} proposed A-INLP approach which dynamically identifies bias-sensitive tokens. Further research in this area helps to mitigate bias in pretrained models and help them to make fair decisions. 

\subsection{Mitigating Fine-Tuning Instabilities} Fine-tuning is the most widely adopted approach to adapt PTLMs to the downstream task. Though fine-tuning achieves good performance, it is unstable i.e., fine-tuning the model with different random seeds results in the large variance of downstream performance. It is believed that catastrophic forgetting and the small size of datasets are possible reasons for fine-tuning instabilities \cite{devlin2019bert,lee2019mixout,dodge2020fine}. However, Mosbach et al. \cite{mosbach2020stability} showed that fine-tuning instability is not caused by any of these two and further showed that fine-tuning instability can be attributed to a) optimization difficulties which lead to vanishing gradients and b) generalization issues. The possible solutions to mitigate fine-tuning instability are a) intermediate fine-tuning \cite{phang2018sentence}  b) mix-out \cite{lee2019mixout} c) smaller learning rates in early epochs and fine-tuning the model for more number of epochs \cite{mosbach2020stability} and d) use of supervised contrastive loss along with cross-entropy loss \cite{gunel2020supervised}. Further work related to this will make fine-tuning more stable.

\section{Conclusion}
In this survey paper, we present a comprehensive review of recent research works in transformer-based pretrained language models. This paper covers various pretraining methods, pretraining tasks, embeddings, downstream adaptation methods, intrinsic and extrinsic benchmarks, useful libraries to work with T-PTLMs. We also present a new taxonomy to categorize various T-PTLMs. We discuss various future research directions which will direct the research community to further improve T-PTLMs.


%



\ifCLASSOPTIONcompsoc
  \section*{Acknowledgments}
\else
  \section*{Acknowledgment}
\fi

Kalyan would like to thank his father Katikapalli Subramanyam for giving a) \$750 to buy a new laptop, 24-inch monitor and study table.  b) \$180 for one year subscription of Medium, Overleaf and Edraw MindMaster software. Edraw MindMaster is used to create all the diagrams in the paper.

\ifCLASSOPTIONcaptionsoff
  \newpage
\fi



\bibliographystyle{IEEEtran}
\bibliography{ammus.bib}
\end{document}